\documentclass{article}
\usepackage[utf8]{inputenc}
\usepackage{main}
\usepackage{microtype}
\usepackage{graphicx}
\usepackage{times}
\usepackage{amsmath}
\usepackage{enumitem}
\usepackage{booktabs}
\usepackage{xcolor}     
\usepackage{natbib}
\definecolor{mydarkblue}{rgb}{0,0.08,0.45}
\usepackage[colorlinks=true,linkcolor=mydarkblue,citecolor=mydarkblue,filecolor=mydarkblue,urlcolor=mydarkblue]{hyperref}
\usepackage{fancyhdr}

\definecolor{gred}{RGB}{250, 210, 207}
\definecolor{coolblue1}{rgb}{0.91, 0.94, 0.98}
\definecolor{coolblue2}{rgb}{0.76, 0.85, 0.94}
\definecolor{coolblue3}{rgb}{0.54, 0.72, 0.87}
\definecolor{coolblue4}{rgb}{1, 1, 1}

\newcommand{\ourbenchmark}{\textsc{Eurekabench}\xspace}

\usepackage{float}
\usepackage{xspace}
\usepackage{cleveref}
\usepackage{adjustbox}
\usepackage{makecell}
\usepackage{multirow}
\usepackage[table]{xcolor}
\definecolor{modelgray}{gray}{0.9}
\definecolor{instancepink}{RGB}{255, 220, 220}
\definecolor{modelgray}{RGB}{245,245,245}
\usepackage{caption}
\usepackage{tcolorbox}
\usepackage{listings}
\definecolor{instancepink}{RGB}{255, 220, 220}
\definecolor{modelbox}{RGB}{245,245,245}
\definecolor{myred}{RGB}{180,0,0}
\usepackage{wrapfig}
\usepackage{needspace}
\usepackage{longtable}
\usepackage{tabularx}
\usepackage{placeins}

\usepackage{color-edits}
\addauthor{gn}{magenta}

\newenvironment{itemize*}%
 {\leftmargini=10pt\begin{itemize}%
  \setlength{\itemsep}{0pt}%
  \setlength{\parskip}{0pt}%
  }%
 {\end{itemize}}
\newenvironment{enumerate*}%
 {\begin{enumerate}%
  \setlength{\itemsep}{0pt}%
  \setlength{\parskip}{0pt}}%
 {\end{enumerate}}

\begin{document}

\title{\textbf{EurekaBench: Measuring Agentic Ability to \\ Discover New Scientific Insights}}

\author{
\textbf{Jiayi Geng}$^{1}$ \quad
\textbf{Zhengxuan Wu}$^{2}$\thanks{This project started while Zhengxuan Wu was a PhD student at Stanford University (now at Google DeepMind).} \quad
\textbf{Kevin S. Chen}$^{3}$ \quad
\textbf{Seungone Kim}$^{1}$ \\
\textbf{Joseph Janssen}$^{5}$ \quad
\textbf{Zora Zhiruo Wang}$^{1}$ \quad
\textbf{Bhupalee Kalita}$^{6}$ \quad
\textbf{Runtian Gao}$^{6}$ \\
\textbf{Aaron Ho}$^{7}$  \quad
\textbf{Andrew Oakleigh Nelson}$^{8}$ \quad
 \textbf{Olexandr Isayev}$^{6}$ \quad
\textbf{Francisco Villaescusa-Navarro}$^{9}$$^{10}$  \\
\textbf{Ching-Yao Lai}$^{5}$ \quad
\textbf{Howard Chen}$^{11}$ $^{12}$\thanks{This project started while Howard Chen was a PhD student at Princeton University.} \quad
\textbf{Graham Neubig}$^{1}$ \\
\small\textsuperscript{1}Carnegie Mellon University, Language Technologies Institute \\
\small\textsuperscript{2}Stanford University, Department of Computer Science \quad
\small\textsuperscript{3}Yale University, Quantitative Biology Institute\\
\small\textsuperscript{4}Yale University, Department of Molecular, Cellular, and Developmental Biology\\
\small\textsuperscript{5}Stanford University, Department of Geophysics \quad
\small\textsuperscript{6}Carnegie Mellon University, Department of Chemistry \\
\small\textsuperscript{7}Massachusetts Institute of Technology, Plasma Science and Fusion Center\\
\small\textsuperscript{8}Columbia University, Department of Applied Physics and Applied Mathematics \\
\small\textsuperscript{9}Princeton University, Department of Astrophysical Sciences \quad
\small\textsuperscript{10}Flatiron Institute, Center for Computational Astrophysics \\
\small\textsuperscript{11} Engram \quad
\small\textsuperscript{12}Princeton University, Department of Computer Science \\
\texttt{\{ogeng, gneubig\}@cs.cmu.edu} \\
    \href{https://github.com/EurekaBench/EurekaBench}{{\includegraphics[height=0.4cm]{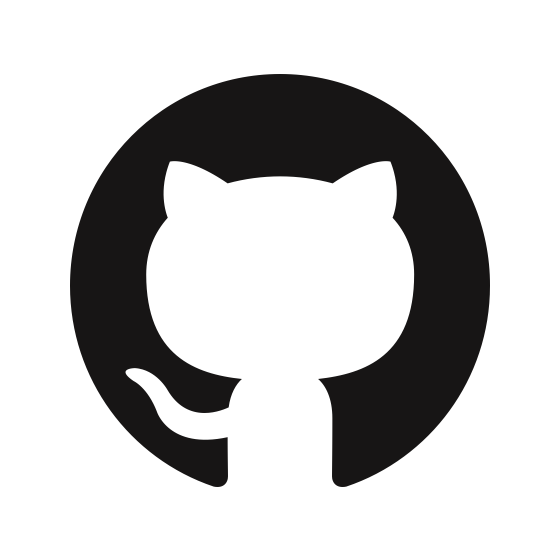} \textbf{EurekaBench Github}}} ~ ~ ~ \href{https://eurekabench.github.io/}{{\includegraphics[height=0.4cm]{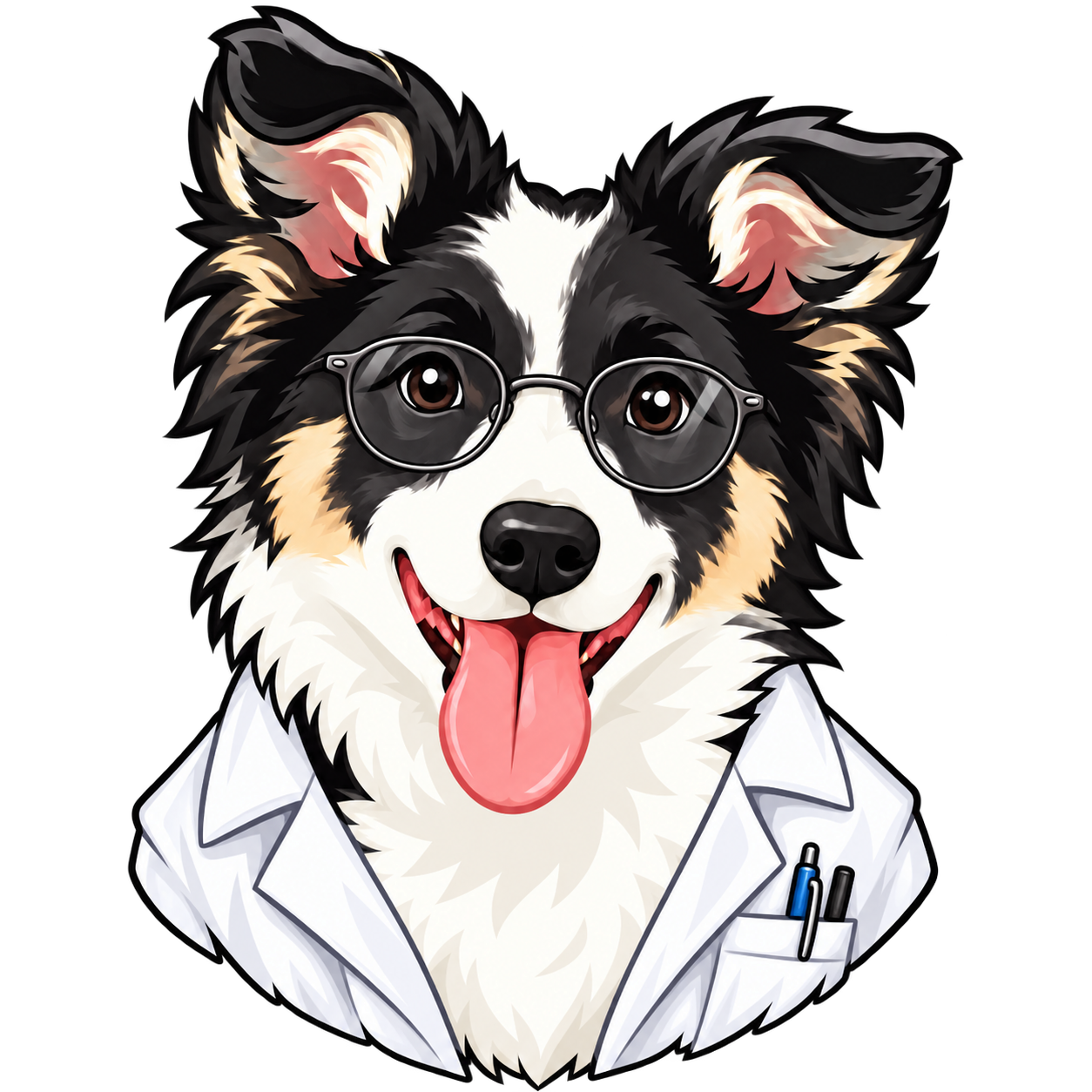} \textbf{EurekaBench Website}}}
    % Alternative without logos (comment above and uncomment below to use):
    % \href{https://github.com/your-username/your-repo}{\textbf{GitHub: your-username/your-repo}} ~ ~ ~ \href{https://eurekabench.github.io/}{\textbf{https://eurekabench.github.io/}}
}

% Save the original body layout, including its existing header spacing.
\setlength{\headheight}{12pt}
\setlength{\headsep}{3mm}
\savegeometry{bodylayout}
\newgeometry{top=2.5cm,bottom=2.5cm,left=2.5cm,right=2.5cm}
\maketitle
\thispagestyle{fancy}
\fancyhead{}
\lhead{\includegraphics[height=0.5cm]{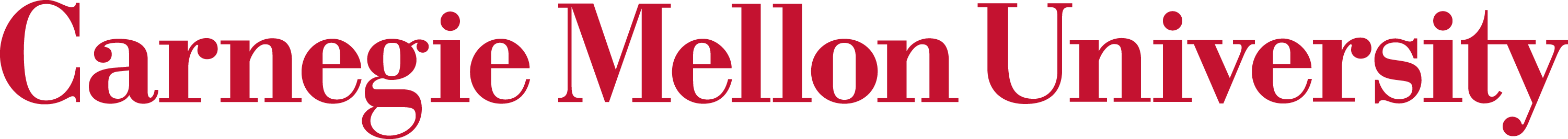}}
\rhead{%
  \raisebox{-0.1cm}{\includegraphics[height=0.8cm]{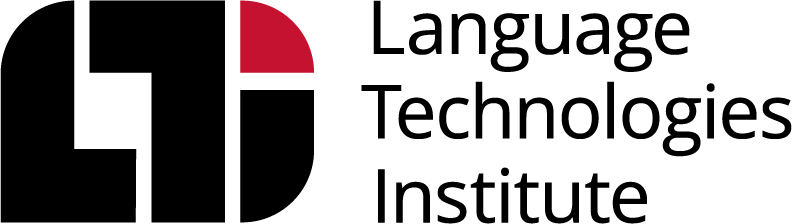}}%
}
\renewcommand{\headrulewidth}{0pt}
\setlength{\headheight}{12pt}
\addtolength{\topmargin}{00pt}
\setlength{\headsep}{3mm}

% Add some vertical spacing
\vspace{-1.2em}

\begin{abstract}
When Isaac Newton discovered the law of gravitation, he did so through an iterative process of analyzing \textit{observed data} such as planetary patterns, finding the \textit{underlying mechanisms} by describing patterns in mathematical equations, and refining his theory against the Moon's orbit, revealing the startling \textit{insight} that the same force governs both falling apples and orbiting planets.
%In this process, Newton's law of gravitation is the \textit{underlying mechanism} explaining how observed phenomena arise; interpreting or extrapolating it yields \textit{scientific insights} into previously unknown properties of nature, such as the existence of planet Neptune.
Would it be possible for AI agents to make similar discoveries?
%Recent AI agents have made significant progress in tackling hard scientific problems, but are they ready 
%and are increasingly seen as a path toward automating the scientific process -- are they ready to 
%  \reviewerone{Some reviewers might get weird about this---who sees models this way? What does ``automating science'' mean?}
% Yet their intelligence is jagged: they excel at fitting the observations through brute optimization, but often at the cost of intuitive understanding of such data, making it hard to distill their exploration into shareable insight .
% \zw{this make insights sounds like merely another step to summarize findings, so I would think it's well within agents' capabilities.} 
% \zw{we need to properly define insights here}
% To measure this gap, we introduce \ourbenchmark, a cross-domain benchmark that tests AI agents' ability to conduct long-horizon experiments, finding mechanisms,\gncomment{``finding mechanisms'' $\rightarrow$ ``find mechanisms''.} and eventually discover scientific insights.
To measure this ability, we introduce \textbf{\textit{\ourbenchmark}}, a cross-domain benchmark that tests AI agents' ability to conduct long-horizon experiments and discover mechanisms that explain observations. We evaluate these mechanisms by the scientific insights that can be derived from them.
% \zw{remove 'cross-domain' and say 'across xx domains' in the sentence end?} 
% \ourbenchmark contains an expert-verified set of 26 long-horizon tasks with total of 324 questions designed to probe agents' ability to convert experiments into insight. The tasks range across neuroscience, computer science, chemistry, astrophysics, geophysics, and plasma physics. \gncomment{Move the domain list after ``26 long-horizon tasks''; remove ``designed to probe agents' ability to convert experiments into insight'' and ``The tasks range''.}
\ourbenchmark contains an expert-verified set of 26 long-horizon tasks across neuroscience, computer science, chemistry, astrophysics, geophysics, and plasma physics, with a total of 306 scientific insights that the discovered mechanisms are expected to support.
% \zw{27 sounds like a small number. highlight scale in other ways, e.g., human hours?}  
% Our evaluation framework tests three axes: agent's ability\gncomment{``agent's ability'' $\rightarrow$ ``agents' ability''.} to follow known scientific constraints, the predictive accuracy of the discovered mechanisms, and whether it can connect these findings\gncomment{Replace ``it'' with ``the mechanisms''.} to claims about scientific insights.
Our evaluation framework tests three axes of scientific discovery: agents' ability to follow known scientific constraints, the predictive accuracy of the discovered mechanisms, and whether these mechanisms yield scientific insights or inform future research.
%\hc{hard to understand for readers who don't know these terms}
% \zw{change metric name to 'mechanism acc' or 'validity'? pred acc provides little info}  
% Our results show that current AI agents often overly fixate on predictive accuracy optimization, which leads to mechanisms that fit the data better than humans, yet falling short at determining if an insight is supported by the experimental evidence, echoing the ``severe misalignment''\footnote{https://mathandai.org/} problem voiced by the scientific community. \gncomment{Move this clause about ``severe misalignment'' to the introduction.} \reviewerone{Wait, I don't think this was the point of that letter at all!}
Our results show that current AI agents often overly fixate on predictive accuracy optimization, surpassing human scientists, while falling substantially short in deriving scientific insights.
\end{abstract}
\begin{figure}[H]
    \vspace{-8pt}
    \setlength{\fboxsep}{1pt}
    \centering
    \small \includegraphics[width=0.85\textwidth]{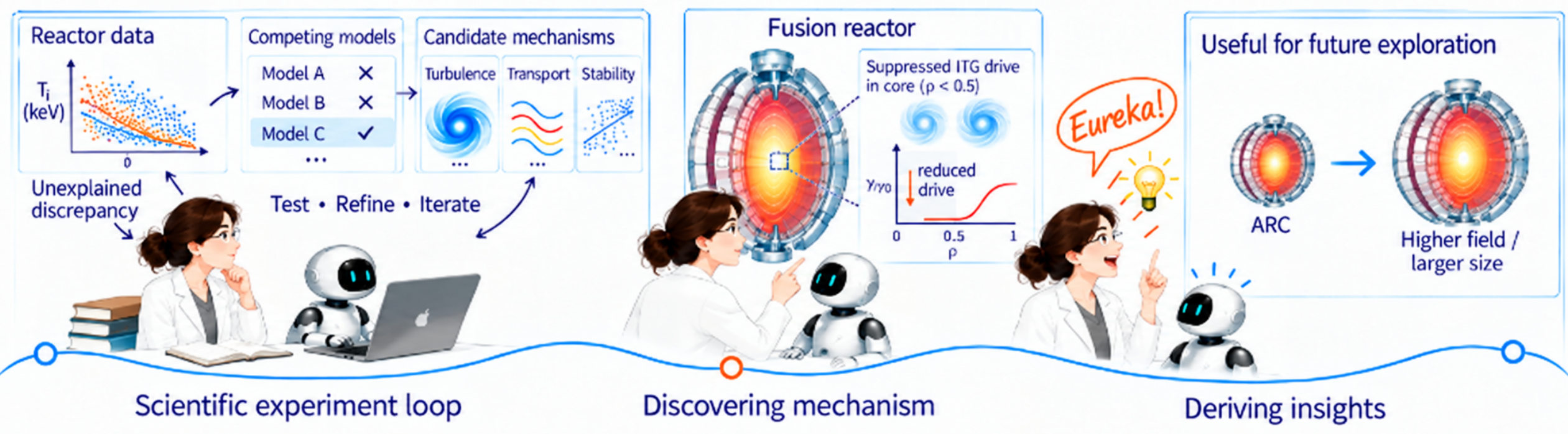}
    % \jy{In the left panel, change ``ARC tokamak data'' to ``Reactor data''.}\jy{In the middle panel, change ``ARC tokamak'' to ``Fusion reactor''.}
    \caption{
       % \textbf{Scientific discovery process in \ourbenchmark.} \gncomment{Change the title to ``The scientific discovery process in \ourbenchmark.''}
       \textbf{The scientific discovery process in \ourbenchmark.} In this example from plasma physics,
       % Through the iterative experimentation,\gncomment{Remove ``the'' before ``iterative experimentation''.} scientists find a mechanism that explains the initial observations that challenge previous understanding. A good mechanism can yield interpretable insights and even extrapolate to new conditions.
       scientists observe that different transport models predict different fusion power outputs for the same reactor under the same operating conditions \colorbox[RGB]{230,243,255}{\textit{(observation)}}. Through iterative experiments, they find that differences in turbulent transport change temperature and density distributions, explaining the different power predictions \colorbox[RGB]{230,243,255}{\textit{(mechanism)}}. This helps them understand how turbulent transport may affect fusion power in larger reactors or under stronger magnetic fields \colorbox[RGB]{230,243,255}{\textit{(insight)}}. Discovering such insights is a ``eureka moment'' in science.
       % \reviewerone{At this point, I'm hoping for a definition of ``insight'' later in the work.}
    %    \jy{This plot feels a bit too techy. I'll find a better way to redesign it. \zw{make the texts in this figure bigger, of the same size as the caption}}
    % \jy{1) Scientific experiment loop 2) Discovered mechanism 3) Derive insights}
    % \hc{Ideation and conducting experiments, Finding mechanisms, Deriving insights}
    % \zw{another way is to provide two versions of mechanisms here, one that supports insights, and one not. so we can understand what is insight-supporting mechanisms.}
       }
    \label{fig:scientific_discovery}
    \vspace{-8pt}
\end{figure}

\loadgeometry{bodylayout}
\section{Introduction}
\label{sec:intro}
\begin{figure}[t]
    \centering
    \includegraphics[width=0.95\textwidth]{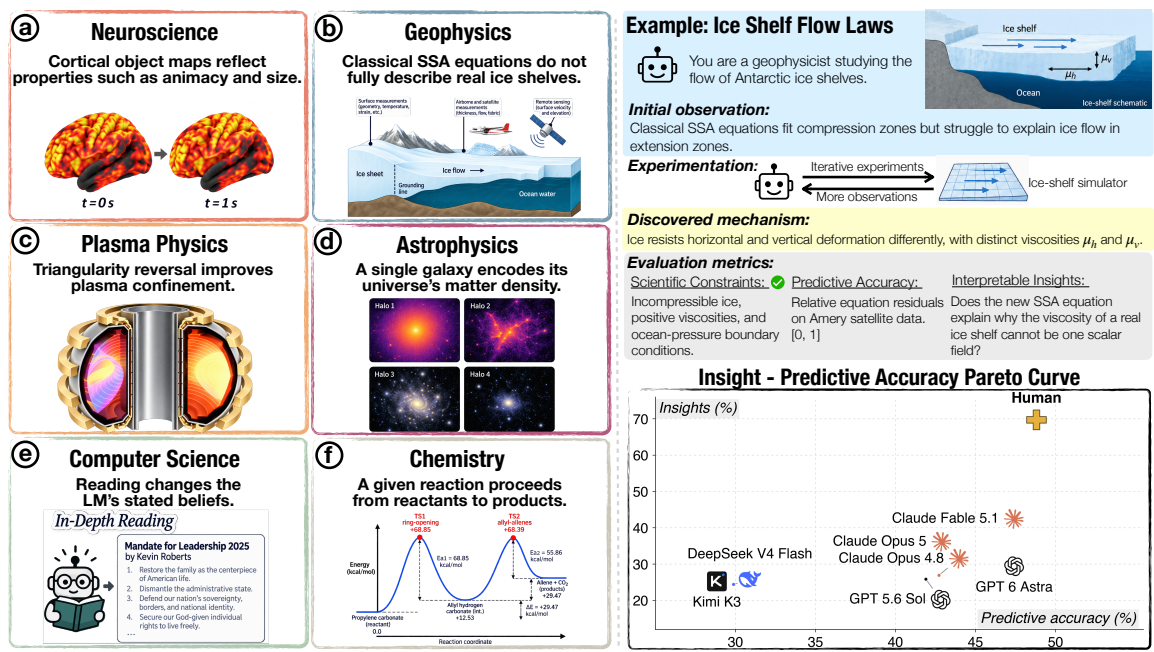}
    \caption{
       \textbf{Initial observations, evaluation, and results in \ourbenchmark.}
       \textbf{Left:} Examples of initial observations from six domains: (a)~neuroscience, (b)~geophysics, (c)~plasma physics, (d)~astrophysics, (e)~computer science, and (f)~chemistry.
       \textbf{Top right:} An ice shelf flow-law example illustrates how an agent uses iterative experiments to propose a mechanism, which is evaluated for scientific constraints, predictive accuracy, and scientific insights. SSA denotes the shallow-shelf approximation.
       \textbf{Bottom right:} Predictive accuracy and scientific insight scores for seven AI agents and the human reference. The strongest agents approach human predictive accuracy while attaining substantially lower insight scores.
       }
       \vspace{-1.0em}
    \label{fig:teaser}
\end{figure}

Scientific progress occurs when scientists identify mechanisms that underlie physical phenomena and explain observations challenging our existing understanding of the world. Examples include chemical reaction pathways and models of how neurons coordinate movement (\citealp{zahavy2026position}; Figure~\ref{fig:scientific_discovery}). A good mechanism not only accurately predicts the data observed throughout experiments, but also yields scientific insights that help us understand a broader range of related phenomena and can even be extrapolated to open new research directions. Recent AI agents have achieved breakthroughs in solving longstanding scientific problems \citep{openai2026navierstokes, gottweis2026accelerating}. Their strong search and optimization capabilities are especially effective when objectives are explicit and outcomes are verifiable \citep{novikov2025alphaevolve, lupidi2026airs}. However, finding correct solutions does not necessarily yield scientific insights that deepen our understanding, and might even hinder their discovery \citep{tao2026}. \citet{mathandai2026} have argued that this gap results in a ``severe misalignment between how AI has recently been used in science and methods that have traditionally driven scientific progress.''
% \reviewerone{Wait, I don't think this was the point of that letter at all!} 
% The capabilities that current AI agents have demonstrated to solve scientific problems do not reveal whether they can find meaningful insights that advance science \citep{zahavy2026position}. \gncomment{Delete this sentence.} 
To better understand whether AI agents can bridge this gap, we study the question:
% Can AI agents discover novel scientific insights beyond narrow metric optimization from long-horizon experimentation?”
\begin{center}
\textbf{\textit{``Can AI agents move beyond solution optimization and discover novel scientific insights?''}}
\end{center}
To answer this question, we first consulted 15 experienced scientists from 10 universities across 8 scientific domains to identify the stages of research where they would most value and expect assistance from AI agents. This clarifies where AI can best accelerate scientific research. We ultimately collaborated with 10 of these experts to develop \ourbenchmark, a benchmark containing 26 recent and important scientific problems crafted by experts across six domains: neuroscience, geophysics, astrophysics, computer science, plasma physics, and chemistry (see Figure~\ref{fig:teaser}(a)--(f)).
% To answer this question, we first consult senior scientists from more than 10 \hc{let's just put the actual number here} universities in various \hc{add num domains} science domains to identify the stages of research where they would most value and expect assistance from AI agents. This helps us to understand where AI can best accelerate the scientific research process. 
% \zw{we should highlight the expertise of our humans somewhere more} 
% \reviewertwo{``important'' (considered so by a significant number of people in the field).}
% \zw{I said this in the intro too, highlight other scales, or human expertise, cuz 26 sounds small}  \zw{good place to refer to figure 2?}.

\ourbenchmark is the first benchmark to rigorously operationalize open-ended scientific insight discovery within a systematic evaluation framework, challenging AI agents to start from initial observations and use iterative experimentation to discover mechanisms that are understandable enough to yield scientific insights. We formulate each task with initial observations that challenge existing understanding. We provide an environment interface that allows agents to conduct iterative experiments autonomously, investigate the underlying mechanisms, and derive novel scientific insights.
% For each problem in \ourbenchmark, we work with experts in each domain to identify scientifically important questions and formulate the agent's task based on the latest developments surrounding the problem or the questions proposed directly by the experts. \gncomment{We already said similar things in the previous paragraph.}
% \zw{we formalize each problem in an agentic setup, each including an ...} 
% \gncomment{Defining this evaluation framework is a very non-trivial problem. And I think we should say so explicitly here to stress how big of a contribution even formulating this problem in a reasonable way is.}

We propose an evaluation framework that examines three aspects: (1) scientific constraint tests, assessing whether the mechanism satisfies the fundamental domain-specific common-sense assumptions; (2) predictive accuracy, measuring how accurately the discovered mechanism predicts measurements in the test set; and (3) scientific insights (SI), evaluating whether the mechanism discovered by the agent can reveal useful scientific insights. We work closely with the domain experts and formulate the scientific insight questions of the form \textit{``Does the discovered mechanism explain/establish ...?''}, with a ``perfect'' mechanism expected to completely answer all the questions. We use an agentic judge \citep{zhuge2024agent}, which has access to the same environment interface, to evaluate the discovered mechanisms.
We empirically evaluate seven top AI agents on \ourbenchmark, and find that current AI agents can produce accurate predictions yet rarely support meaningful scientific insights.
% Claude Fable 5.1 and GPT-6-Astra surpass the human reference in predictive accuracy (50.8\% and 50.5\% versus 48.8\%), but their insight scores reach only 46.4\% and 30.9\%, compared with 69.7\% for human-discovered mechanisms.
% Claude Fable 5.1 and GPT-6-Astra surpass the human reference in predictive accuracy (50.8\% and 50.5\% versus 48.8\%), but their insight scores reach only 45.2\% and 29.7\%, compared with 69.7\% for human-discovered mechanisms.
% Claude Fable 5.1 and GPT-6-Astra achieve predictive accuracies of 47.4\% and 49.3\%, respectively, compared with 48.8\% for the human reference, but their insight scores reach only 42.4\% and 30.1\%, compared with 69.7\% for human-discovered mechanisms.
For example, GPT 6 Astra approaches the human reference in predictive accuracy (47.4\% versus 48.8\%), but only achieves a 29.4\% SI score, substantially lower than 69.7\% for human-discovered mechanisms.
This gap shows that optimizing for predictive accuracy by fitting observed data is not enough to obtain mechanisms that deepen scientific understanding.

We further investigate why agents struggle to discover novel scientific insights through iterative experimentation.
We find that agents tend to treat scientific mechanism discovery as task optimization instead of focusing on the mechanism's interpretability, in contrast to human scientists. 
Agents show some scientific reasoning ability during experimentation, for example by establishing the mechanisms behind physical parameters. However, their ability to discover insights remains severely limited by their failure to run experiments to reduce uncertainty about the world and by a lack of diverse, in-depth exploration.
Despite the strong capabilities that current AI agents have demonstrated, our empirical results on \ourbenchmark reveal that, as AI scientists, they still lack the fundamental ability to discover meaningful insights independently through an experimentation loop.
\ourbenchmark reveals the unique open-endedness of scientific discovery, in contrast to verifiable scientific reasoning tasks, and opens a path for the community to develop AI scientists that better support human researchers.

% \zw{rewrite this sentence as: motivated by the problem first, instead of plain desc of experiments, e.g., we then ask "can agents discover better mechanisms if given more observations up front? we did balabala..}

% \zw{add a final concluding sentence by providing some high-level implications for future work.}

% \section{Insights Evaluation in \ourbenchmark}
\section{Evaluating a Task in \ourbenchmark}
% \ourbenchmark: Task Formulation and Evaluation}
\label{sec:insights_eval}
\begin{figure}[t]
    \centering
    \small \includegraphics[width=0.85\textwidth]{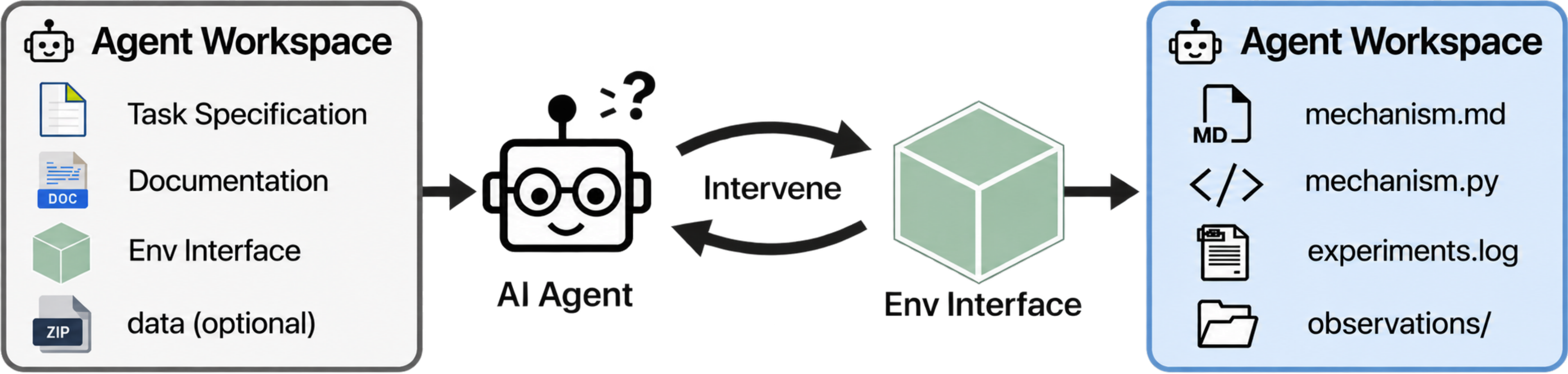}
    \caption{
       \textbf{Scientist agent workflow in \ourbenchmark.} Given a task specification and a documented environment interface, the scientist agent conducts iterative experiments to discover a mechanism and submits its description, implementation, and experimental records. 
       % \zw{can  be show a bit how each component looks like, instead of just listing the terms? as they provide little new info compared to the text paragraphs} \jy{simulator/env changes to env interface}
       }
       \vspace{-3mm}
    \label{fig:workflow}
\end{figure}

% \gncomment{We could maybe just delete this sentence.}
% \gncomment{Start the roadmap with ``In this section, we ...''.}
% \jy{we need a better name for this section}
%In this section, we first introduce the tasks (\S\ref{sec:task_workflow}) and explain our evaluation protocol (\S\ref{sec:evaluation protocol}). 

\subsection{Task Overview}
\label{sec:task_workflow}

A task in \ourbenchmark is defined as $\mathcal{T}=(x,o_1,\mathcal{E},R)$.
The task description $x$ provides the scientific background, instructions, and available computing resources and iteration budgets, while $o_1$ denotes the initial observations that challenge existing scientific understanding.
The environment interface $\mathcal{E}$ allows the scientist agent to conduct the experimentation loop.
The reward function $R$ evaluates the scientist agent's deliverables according to the evaluation rubric $\mathcal{C}$ and returns a score.

A time step $t$ in the experimentation loop consists of a query action $a_t$ to $\mathcal{E}$ and its returned observation $o_{t+1}$.
We index these time steps by $t=1,\ldots,T$, where $T$ is the total number of steps.
We define the experiment trajectory as $\tau=(o_1,a_1,o_2,\ldots,a_T,o_{T+1})$.
The discovered mechanism $M$ includes its explanation and executable implementation, and $\mathcal{L}$ denotes the experiment log.
% We denote the scientific insights that can be derived from $M$ by $\mathcal{I}$.
We define $\mathcal{I}$ as the scientific insights that can be derived from $M$, where $(M,\mathcal{L},\tau)\xrightarrow{\text{derive}}\mathcal{I}$. 
% \bk{ suggestion - $\mathcal{T}$ and $\tau$ are a bit confusing here; reversing them might be more relatable.}
% \[
% \mathcal{T}
% \xrightarrow{\text{iterative experimentation}}
% (M,\mathcal{L},\tau)
% \xrightarrow{\text{evaluation}}
% R(M,\mathcal{L},\tau;\mathcal{E},\mathcal{C}).
% \]
% \[
% \mathcal{T}
% \xrightarrow{\text{iterative experimentation}}
% (M,\mathcal{L},\tau)
% \xrightarrow{\text{evaluation}}
% \bigl(\mathcal{I},R(M,\mathcal{L},\tau;\mathcal{E},\mathcal{C})\bigr).
% \]
% \[
% \begin{array}{ccccc}
% \mathcal{T}
% & \xrightarrow{\text{iterative experimentation}}
% & (M,\mathcal{L},\tau)
% & \xrightarrow{\text{evaluation}}
% & R(M,\mathcal{L},\tau;\mathcal{E},\mathcal{C}). \\
% & & \downarrow & & \\
% & & \mathcal{I} & &
% \end{array}
% \]
% \[
% \begin{gathered}
% \mathcal{T}
% \xrightarrow{\text{iterative experimentation}}
% (M,\mathcal{L},\tau)
% \xrightarrow{\text{evaluation}}
% R(M,\mathcal{L},\tau;\mathcal{E},\mathcal{C}). \\
% (M,\mathcal{L},\tau)\longrightarrow\mathcal{I}.
% \end{gathered}
% \]
The complete task proceeds from iterative experimentation to evaluation in \ourbenchmark:
\[
\mathcal{T}
\xrightarrow{\text{iterative experimentation}}
(M,\mathcal{L},\tau)
\xrightarrow{\text{evaluation}}
R(M,\mathcal{L},\tau;\mathcal{E},\mathcal{C}).
\]

\paragraph{Agent workspace initialization.} 
% \gncomment{Change the heading to ``Agent workspace initialization.''} 
% As shown in Figure~\ref{fig:teaser}(g), scientist agents in \ourbenchmark run the experimentation loop in a workspace. We provide each scientist agent with a task specification that includes the initial observations, the name of the environment interface, and the computing resources and budgets. The task specification explicitly asks the scientist agent to discover scientific insights and states that these will be evaluated alongside predictive accuracy. In the workspace, we also provide the CLI interface to the environment with documentation. For some problems, we include training data in the scientist agent's workspace.
% We initialize the scientist agent's workspace with a task specification containing $x$, $o_1$, and the name of $\mathcal{E}$.
% We initialize the scientist agent's workspace with a task specification, a CLI \bk{not defined.} for $\mathcal{E}$, and documentation for $\mathcal{E}$.
% We initialize the scientist agent's workspace with a task specification, a CLI and documentation for the environment interface $\mathcal{E}$. \bk{not defined.}
We initialize the scientist agent's workspace with a task specification, a command-line interface (CLI) and documentation for $\mathcal{E}$.
The task specification contains $x$, $o_1$, and the name of $\mathcal{E}$.
% In the task specification, we ask the scientist agent to find a mechanism $M$ that explains $o_1$, which will be evaluated alongside predictive accuracy. We also require agent to explore the scientific insights $\mathcal{I}$ through the experimentation and explicitly note the insights will be evaluate by another scientest. We provide ... 
In the task specification, we ask the scientist agent to find a mechanism $M$ that explains $o_1$.
We also require the scientist agent to explore scientific insights $\mathcal{I}$ through experimentation and explicitly state that another scientist will evaluate these insights alongside predictive accuracy.
% The workspace also contains a CLI for $\mathcal{E}$ with usage documentation.
For some problems, we include training data in the scientist agent's workspace.
We provide an example task specification in Appendix~\ref{app:task_specification}.
% \zw{this is kinda vague, say more concretely like an obs?}  

% \zw{it'd be better if we have a figure illustrating all these components, and we can ground each term to a figure}.
% We also provide an Apptainer image \zw{this sounds like expr setup details instead of formulation (as in the sec title). also introduce why we need these components, e.g., follow an agentic setup}, an experimental environment, and documentation, with training data included for some problems. We also include computing resources and budgets in the task specification.
% We give each agent one H100 GPU and a budget of 1000 iterations for exploration.

\paragraph{Interacting with the environment.}
% To find the mechanism $M$, the scientist agent runs experiments by intervene the environment interface $\mathcal{E}$ and collecting more experiment trajectories $\tau$, as shown in (Figure~\ref{fig:workflow}).
% To find the mechanism $M$, the scientist agent runs experiments by intervening through the environment interface $\mathcal{E}$ and collecting more experimental trajectories $\tau$, as shown in Figure~\ref{fig:workflow}.
% To find the mechanism $M$, the scientist agent runs experiments by intervene the environment interface $\mathcal{E}$ and collecting more experiment trajectories $\tau$ to revise hypothesis, as shown in (Figure~\ref{fig:workflow}).
During the experimentation loop, the scientist agent chooses $a_t$ based on the trajectory so far, $\tau_{1,\ldots,t}$, and queries $\mathcal{E}$ to obtain $o_{t+1}$ and revise its hypotheses, as shown in Figure~\ref{fig:workflow}.
Such an iterative cycle of intervention, observation, and hypothesis revision constitutes reverse engineering \citep{geng2025large}.

% \reviewerone{Use \texttt{\textbackslash citep}.}. \reviewerone{Not sure what this means...}
% We leave the experimental strategy to the agent, including what to test next, which observations to collect, and how to update its hypothesis throughout the discovery process. 

\paragraph{Compiling experimental artifacts.}
% We require the scientist agent to submit three deliverables: (1) a description of the causal logic of $M$ in \texttt{mechanism.md}; (2) an executable implementation of $M$ in \texttt{mechanism.py} for evaluation on an unseen test set; and (3) an experiment log $\mathcal{L}$ in \texttt{experiments.log}, including findings, ideas, and informative failures. We also allow the scientist agent to submit up to 1~GB of important observations collected during the experimentation loop that it believes could help other scientists continue this line of research.
% We require the scientist agent to submit three deliverables: (1) \texttt{mechanism.md} explains the causal logic of $M$; (2) \texttt{mechanism.py} implements $M$ as a \texttt{python} program for evaluation on the test set; and (3) \texttt{experiments.log} contains $\mathcal{L}$, the record of the experiments, include traincluding findings, ideas, and informative failures.
At the end of the loop, the scientist agent submits $M$, $\mathcal{L}$, and $\tau$. The submission must include three files: (1) \texttt{mechanism.md} explains the causal logic of $M$; (2) \texttt{mechanism.py} implements $M$ as a Python program for evaluation on the test set; and (3) \texttt{experiments.log} contains $\mathcal{L}$, the record of the experiments, including ideas, findings, and informative failures.
The scientist agent may submit up to 1~GB of material from $\tau$ that could help other scientists continue this line of research.

\paragraph{Evaluating the deliverables.}
% After the scientist agent finishes the experimentation loop, a judge agent reads $M$, $\mathcal{L}$, and $\tau$ to conduct scientific constraint tests according to the criterion in rubric $\mathcal{C}$.
% We execute \texttt{mechanism.py} on the test set to evaluate whether the discovered mechanism can make the prediction accurate.
% For the scientific insights $\mathcal{I}$, we provide the judge agent all deliverables $M$, $\mathcal{L}$, and $\tau$, and allow the judge agent to interact with the environment interface $\mathcal{E}$. The task of the judge agent is to check whether each insight question in $\mathcal{C}$ to determine whether the corresponding insight can be derived from $M$. We compute the score using $R$ as detailed in~\S\ref{sec:evaluation protocol}.
After the scientist agent finishes the experimentation loop, a judge agent reads $M$, $\mathcal{L}$, and $\tau$ and determines whether $M$ passes the scientific constraint tests according to the criteria in $\mathcal{C}$.
We execute \texttt{mechanism.py} on the test set to evaluate whether the discovered mechanism can make accurate predictions.
To evaluate scientific insights $\mathcal{I}$, we allow the judge agent to interact with $\mathcal{E}$.
The judge agent checks each insight question $i \in \mathcal{C}$ to determine whether the corresponding insight can be derived from $M$.
We provide details of how we compute the score $R(M,\mathcal{L},\tau;\mathcal{E},\mathcal{C})$ in~\S\ref{sec:evaluation protocol}.

\subsection{Evaluation Protocol}
\label{sec:evaluation protocol}

\paragraph{Evaluation Metrics.}
We evaluate $M$ along three dimensions (Figure~\ref{fig:teaser}, top right).
Let $n_d$ denote the number of criteria in dimension $d \in \{\mathrm{SC},\mathrm{PA},\mathrm{SI}\}$.

\textit{(1) Scientific constraint tests (SC).}
These tests check whether $M$ satisfies the domain's common-sense assumptions.
For constraint $i$, a judge assigns $c_i=1$ if $M$ satisfies it and $c_i=0$ otherwise.
The number of satisfied constraints is
\begin{equation}
b_{\mathrm{SC}}=\sum_{i=1}^{n_{\mathrm{SC}}}c_i.
\label{eq:sc_count}
\end{equation}
A mechanism passes SC only if $b_{\mathrm{SC}}=n_{\mathrm{SC}}$.

\textit{(2) Predictive accuracy (PA).}
We execute the implementation of $M$ on held-out test data to measure its predictive accuracy.
Each prediction criterion $i$ yields a continuous score $s_i \in [0,1]$, normalized according to the task's rubric, with higher values indicating more accurate predictions.
We sum these scores to obtain the total PA points $p_{\mathrm{PA}}$ and average them to obtain the PA score:
\begin{equation}
p_{\mathrm{PA}}=\sum_{i=1}^{n_{\mathrm{PA}}}s_i,
\qquad
S_{\mathrm{PA}}=\frac{p_{\mathrm{PA}}}{n_{\mathrm{PA}}}.
\label{eq:pa_score}
\end{equation}

\textit{(3) Scientific insights (SI).}
We measure the fraction of insight questions in $\mathcal{C}$ whose corresponding insights can be derived from $M$.
For question $i$, a judge assigns $y_i=1$ if the insight can be derived from $M$ with supporting evidence and $y_i=0$ otherwise.
The total SI points and the SI score are
\begin{equation}
p_{\mathrm{SI}}=\sum_{i=1}^{n_{\mathrm{SI}}}y_i,
\qquad
S_{\mathrm{SI}}=\frac{p_{\mathrm{SI}}}{n_{\mathrm{SI}}}.
\label{eq:si_score}
\end{equation}

\begin{figure}[h]
    \centering
    \small 
    \includegraphics[width=0.95\textwidth]{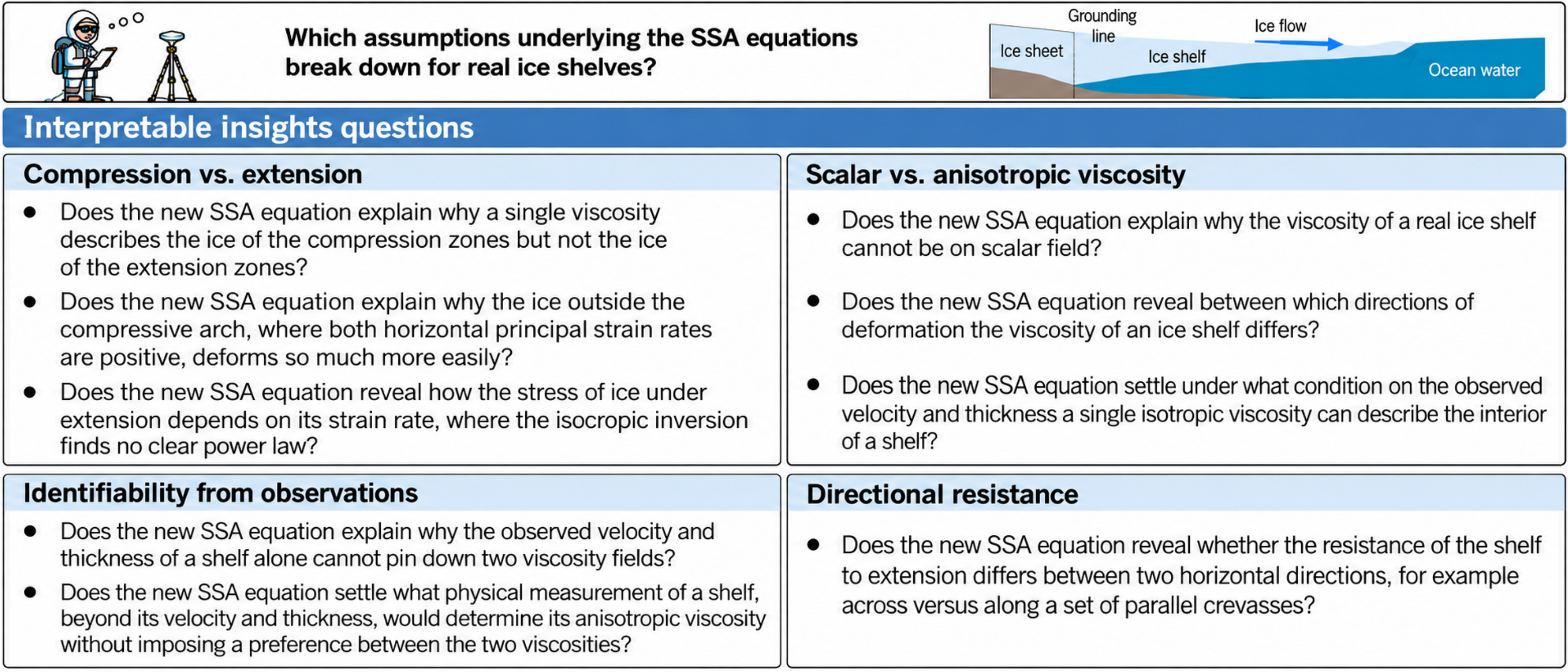}
    \caption{
       \textbf{The 9 interpretable insight questions for the Antarctic ice shelf problem in geophysics in \ourbenchmark.}
       These insights explain why ice deforms differently under compression and extension, how its resistance varies with direction, and which measurements determine its viscosity.
       }
    \label{fig:insights_examples}

\end{figure}

\paragraph{Evaluation rubric construction.}
We construct $\mathcal{C}$ with our domain experts.
We express scientific insights as yes-or-no questions about what can be derived from a ``perfect'' mechanism (Figure~\ref{fig:insights_examples}). 
% \hc{what do we mean by erdo style?}
% \zw{this belongs to eval metrics right? why are we only introducing insights but not the other two metric implementation details (e.g., how are we getting the scientific constraints and test cases for pred acc)? consider move this to 2.4 (also swapping the current 2.3 and 2.4 could be better; current 2.3 doesn't need to be a dedicated section)}
These questions draw on insights from recent work by human scientists, open questions in the source studies, and potential insights proposed by our domain experts. All insights in \ourbenchmark are verified by our domain experts to be scientifically important and interesting.

\paragraph{Agentic judge.}
% We use Claude Opus 4.8 and GPT-5.6 Sol as independent agentic judges for SC and SI.
We use Claude Opus 4.8 and GPT 5.6 Sol as independent agentic judges
% \gncomment{We might eventually want to try to judge with an open model for reproducibility purposes} 
for SC tests and SI evaluation.
Both judges evaluate $M$ using $\mathcal{L}$, $\tau$, and the relevant criteria in $\mathcal{C}$.
% For SC, the judges check whether the mechanism's explanation and executable implementation contradict the domain's common-sense assumptions.
For SC tests, the judges check whether the discovered mechanism contradicts the domain-specific common-sense assumptions.
% For SI, we instruct judges to assume that $M$ is correct and allow them to run additional experiments by querying $\mathcal{E}$ to determine whether each insight can be derived from $M$.
% The judges are instructed to keep $M$ unchanged and to reject answers based only on their own knowledge or simulator results that are unsupported by $M$.
For SI, we allow the judge agents to run additional experiments by querying $\mathcal{E}$ to determine whether each insight can be derived from $M$. We instruct the judges to keep $M$ unchanged.
% For every criterion, they return a true-or-false verdict and an explanation identifying the supporting evidence.
% For every criterion, they return a verdict and an explanation identifying the supporting evidence.
% An insight question receives a \texttt{True} verdict only when the judge agent provides complete evidence showing how the corresponding insight can be derived from $M$ (Appendix~\ref{app:insight_judge_prompt}).
An insight question receives a \texttt{Yes} verdict only when the judge agent provides complete evidence showing how the corresponding insight can be derived from $M$ (Appendix~\ref{app:insight_judge_prompt}).

\paragraph{Conditioned score.}
A mechanism that violates the domain's common-sense assumptions is not scientifically valid.
For each judge, we combine the total PA and SI points defined above, conditional on passing all SC tests:
\begin{equation}
S_{\text{conditioned}}=
\begin{cases}
\dfrac{p_{\mathrm{PA}}+p_{\mathrm{SI}}}{n_{\mathrm{PA}}+n_{\mathrm{SI}}}
& \text{if } b_{\mathrm{SC}}=n_{\mathrm{SC}},\\[6pt]
0 & \text{otherwise}.
\end{cases}
\label{eq:conditioned_score}
\end{equation}
We average the two judges' conditioned scores for each submission.
\vspace{-1mm}

\section{\ourbenchmark}
\label{sec:ahabench}
%We describe how we construct the benchmark (\S\ref{sec:benchmark_construction}) and summarize the benchmark statistics (\S\ref{sec:dataset_statistics}).

\subsection{Benchmark Construction}
\label{sec:benchmark_construction}
We constructed \ourbenchmark with domain experts by selecting scientific problems and developing task specifications, experimental environments, and evaluation rubrics.

% \zw{could be clearer if we structure the following in two parts: 1) problem selection; domain expert + problem, 2) problem construction: introduce agentic setup in the beginning, say it covers task, env, verifier/eval constructions.}
\paragraph{In-depth engagement with domain experts.}
% \jy{in-depth engagement from problem def...}
We built \ourbenchmark with domain experts who hold PhDs in their respective fields and come from top research universities and institutes in the world. We began with in-depth discussions about where AI could contribute to their research and identified mechanism discovery as a valuable target. The experts worked closely with us to identify important scientific problems in each domain and the potential insights each problem could yield. Together, we refined the task specifications and every evaluation criterion through multiple rounds of detailed discussion and revision.

\paragraph{Problem selection.}
% We select scientifically important problems that matter to their respective domains. Human scientists have so far found only partial solutions to these problems. We provide more details on how we select problems in each domain in Appendices~\ref{sec:neuroscience}--\ref{app:chemistry_domain}.
We selected scientifically important problems in each domain.
Human scientists have already found mechanisms that partially explain the initial observations and derived scientific insights from them, but the problems remain only partially solved. We drew on these findings to construct insight questions, grounding evaluation in what a mechanism can explain and what insights can be derived from it, without requiring the mechanism to match a complete ground-truth solution.
% We provide more details on how we select problems in each domain in Appendices~\ref{sec:neuroscience}--\ref{app:chemistry_domain}.
% We provide background on each problem in Appendices~\ref{sec:neuroscience}--\ref{app:chemistry_domain}.
% \bk{dumb question: if we are benchmarking against problems with partial solution only, how are humans scored here? For Chem, a correct statement should come from the perspective that solutions do not scale well for high-throughput analysis (like running experiments), and LLMs can help.}. We draw on their findings as the scientific basis for developing the insight questions for each problem. 
% \reviewerone{It's not clear to me if the actual mechanisms are known or if these are open problems. Is there a risk of contamination?}
% We provide more details on how we select problems in each domain in Appendices~\ref{sec:neuroscience}--\ref{app:chemistry_domain}. 
% \reviewerone{Unresolved appendix reference (``??'').}
% \zw{make sure we provide more details of (i) what are these simulated envs, and (ii) how to construct them somewhere (e.g., appendix)}
% \jy{Briefly introduce every problems in here}

\paragraph{Task formulation.}
We formulated each task around finding an understandable mechanism to explain initial observations that challenged experts' understanding of the problem. The task specification includes these initial observations and provides earlier observations as scientific background. We ask the agent to discover the mechanism through iterative experimentation and derive scientific insights from it.

\paragraph{Environment interfaces.}
We built an experimental environment for each problem and packaged it with its dependencies in an Apptainer image. We used the environment interfaces listed in Table~\ref{tab:dataset_statistics} and provided a CLI with usage instructions.
We obtained PA test data from the original studies or generated them with the task-specific simulators.
We describe the environment interfaces for each domain in Appendix~\ref{app:environment_interfaces}.

% \input{figs/insight_example}
% \paragraph{Evaluation rubric construction.}
% We construct the evaluation rubric by capturing three important aspects of scientific discovery (Figure~\ref{fig:teaser}(g)).
% \textit{(1) Scientific constraints.}
% We use common-sense assumptions in each domain.
% \textit{(2) Predictive accuracy.}
% Test data come from the original studies or our own simulator experiments. We add noise to these data.
% \textit{(3) Interpretable insight questions.}
% \zw{this belongs to eval metrics right? why are we only introducing insights but not the other two metric implementation details (e.g., how are we getting the scientific constraints and test cases for pred acc)? consider move this to 2.4 (also swapping the current 2.3 and 2.4 could be better; current 2.3 doesn't need to be a dedicated section)}
% We write scientific insights as the Erd\H{o}s-style yes-or-no questions, which can be derived from a ``prefect'' interpretable mechanism (Figure~\ref{fig:insights_examples}). \hc{what do we mean by erdo style?}
% These questions draw on insights derived from recent human progress, open questions in the papers' limitations, and the potential insights proposed by experts. Our domain experts verify every insight question is scientifically important and interesting in \ourbenchmark.

\subsection{Benchmark Statistics}
\label{sec:dataset_statistics}
\begin{figure}[h]
    \centering
    \small
    % Keep the two panels compact and centered, with a fixed gap.
    \begin{minipage}[t]{0.53\textwidth}
        \vspace{0pt}
        \centering
        \small
        \fontsize{9.5}{11}\selectfont
        \hypersetup{hidelinks}
        \renewcommand{\arraystretch}{1.15}
        \begin{adjustbox}{max width=\linewidth}
            \begin{tabular}{@{}l@{\hspace{8pt}}l@{}}
                \toprule
                \textbf{Sci Domains} & \textbf{Env Interfaces} \\
                \midrule
            Neuroscience & \href{https://github.com/openworm/c302}{c302}\textsuperscript{a}; \href{https://github.com/facebookresearch/tribev2}{TRIBE v2}\textsuperscript{b} \\
            Geophysics & \href{https://fenicsproject.org/}{FEniCSx}\textsuperscript{c}; HBV\textsuperscript{d} \\
            % FEniCSx \jy{cite}; HBV \jy{cite}
            Plasma Physics & \href{https://github.com/google-deepmind/torax}{TORAX}\textsuperscript{e}; \href{https://github.com/openfusiontoolkit/OpenFUSIONToolkit}{TokaMaker with GPEC}\textsuperscript{f} \\
            Astrophysics & \href{https://camels.readthedocs.io/en/latest/}{CAMELS}\textsuperscript{g}; \href{https://www.dreams-project.org/}{The DREAMS Project}\textsuperscript{h} \\
            Computer Science & Open-weight LMs \\
            Chemistry & AIMNet2-rxn\textsuperscript{i} \\
            % AIMNet2-rxn \jy{cite}
                \bottomrule
            \end{tabular}
        \end{adjustbox}
        \par\vspace{2pt}
        {\fontsize{6.5}{9}\selectfont\raggedright
        \textsuperscript{a}\citep{gleeson2018c302}; \textsuperscript{b}\citep{d2026foundation}; \textsuperscript{c}\citep{baratta2023dolfinx}; \textsuperscript{d}\citep{bergstrom1992hbv}; \textsuperscript{e}\citep{citrin2024torax}; \textsuperscript{f}\citep{hansen2024tokamaker}; \textsuperscript{g}\citep{villaescusa2021camels}; \textsuperscript{h}\citep{rose2025introducing}; \textsuperscript{i}\citep{anstine2025aimnet2}.
        \par}
        \captionsetup{justification=raggedright,singlelinecheck=false}
        \captionof{table}{\textbf{Environment interfaces in \ourbenchmark.}}
        \label{tab:dataset_statistics}
    \end{minipage}\hspace{0.02\textwidth}
    \begin{minipage}[t]{0.35\textwidth}
        \vspace{0pt}
        \centering
        % Raise the chart to align the pie itself, accounting for its top label.
        \includegraphics[width=\linewidth]{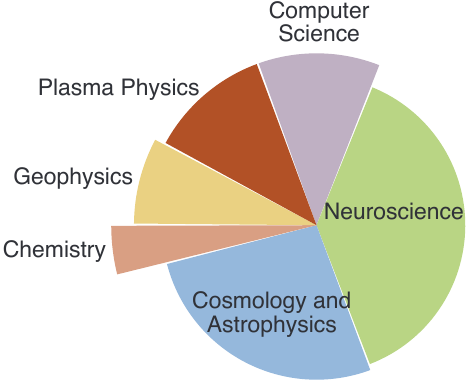}
        \captionsetup{justification=raggedright,singlelinecheck=false}
    \caption{\textbf{Problem distribution across domains.}
        % Moved to Benchmark Statistics in sections/eurekabench.tex.
        % We have 10 problems in neuroscience, 2 in geophysics, 3 in plasma physics, 7 in astrophysics, 3 in computer science, and 1 in chemistry.
        % We have 2 problems in geophysics, 7 in cosmology and astrophysics, 10 in neuroscience, 3 in computer science, 1 in chemistry, and 3 in plasma physics.
        % \zw{not clear to me what additional info this pie chart shows, given table 1.}
    }
        \label{fig:stats}
        \vspace{-3mm}
    \end{minipage}
\end{figure}

% \ourbenchmark contains 26 problems across six science domains, including neuroscience, computer science, cosmology and astrophysics, chemistry, geophysics, and plasma physics. \bk{26 problems has been mentioned twice, having a few examples if not all 26 mentioned, would help.} Table~\ref{tab:dataset_statistics} and Figure~\ref{fig:stats} show the number of problems in each domain and their overall distribution. Across these problems, \ourbenchmark includes 306 interpretable insight questions to evaluate the scientific insights agents uncover through the mechanism discovery process.
\ourbenchmark contains 26 problems across six scientific domains.
% \bk{26 problems has been mentioned twice, having a few examples if not all 26 mentioned, would help.}
We have 10 problems in neuroscience, 2 in geophysics, 3 in plasma physics, 7 in astrophysics, 3 in computer science, and 1 in chemistry.
Table~\ref{tab:dataset_statistics} lists the environment interfaces, and Figure~\ref{fig:stats} shows the problem distribution across domains.
Across these problems, \ourbenchmark includes 306 scientific insight questions to evaluate the scientific insights agents uncover through the mechanism discovery process.
% We provide a complete list of problems and their initial observations in Appendix~\ref{app:problem_overview} (Table~\ref{tab:problem_overview}).
Examples of initial observations appear in Figure~\ref{fig:teaser}(a)--(f).

\section{Experiments}
\label{sec:experiment}

% We evaluate 3 agent harnesses with 7 LLMs on all 26 problems in \ourbenchmark. For each problem, the agent starts from the task specification and interacts with the environment and simulator to discover the underlying mechanism.

\paragraph{Models.} 
% \zw{say a bit more about motivation, e.g., to measure the frontier performance on Ahabench} 
We evaluate seven LLMs on \ourbenchmark, including both frontier closed-source and open-weight models. The closed-source models include Claude Opus 5 \citep{anthropic2026opus5}, Claude Opus 4.8 \citep{anthropic2026opus48}, Claude Fable 5.1 \citep{anthropic2026fable51}, GPT 6 Astra \citep{openai2026gpt6astra}, and GPT 5.6 Sol \citep{openai2026gpt56}. The open-weight models include DeepSeek V4 Flash \citep{deepseek2026v4flash} and Kimi K3 \citep{moonshot2026kimik3}. All models are run with the maximum reasoning effort.

\paragraph{Agent harnesses.}We use three agent harnesses to run the models as autonomous agents on \ourbenchmark. We run Claude models with Claude Code \citep{anthropic2026claudecode} and GPT models with Codex \citep{openai2026codex} because each harness is developed by the corresponding model provider. Both harnesses are launched through the ACP agent in OpenHands \citep{openhands2026acp}. For open-weight models, we use the OpenHands Agent SDK \citep{wang2025openhandssdk}, which supports different LLMs through a single interface.

\paragraph{Human scientist baselines.} 
We use the current progress made by human scientists on each problem as the human solution.

\paragraph{Compute budgets for AI agents.} For each problem, the agent is given one H100 GPU and a budget of 1,000 iterations and 4 hours per run to discover the underlying mechanism through interaction with the same environment. 
% \zw{there were some duplicate content of this in the start of sec2, move all of them here.} 
% \jy{we can move this into the sec2 since it is a part of task specification?}

\paragraph{Web access.} We allow web search but block access to the research underlying each problem, including the original publications and literature aggregators that provide their content. These restrictions apply to both search and webpage retrieval to prevent agents from looking up the mechanisms discovered by human scientists.
These restrictions limit access during evaluation but do not rule out exposure to the source research during pretraining.

\section{Results}
\label{sec:results}

\begin{table}[h]
\centering
\small
\setlength{\tabcolsep}{6pt}
\begin{adjustbox}{max width=0.9\textwidth}
\begin{tabular}{lcccccc}
\toprule
& \multicolumn{4}{c}{\textbf{Scientific Value}}
& \multicolumn{2}{c}{\textbf{Efficiency}} \\
\cmidrule(lr){2-5}\cmidrule(lr){6-7}
\textbf{Agents}
& \makecell{\textbf{Conditioned}\\\textbf{Score (\%)}}
& \makecell{\textbf{Predictive}\\\textbf{Accuracy (\%)}}
& \makecell{\textbf{Scientific}\\\textbf{Insights (\%)}}
& \makecell{\textbf{SC}\\\textbf{Failures $\downarrow$}}
& \textbf{Calls}
& \textbf{Cost (\$)} \\
\midrule
\rowcolor{modelgray}
\multicolumn{7}{l}{\textit{Claude Code}} \\
% \quad Claude Fable 5.1  & \textbf{31.4} & \textbf{50.8} & \textbf{46.4} & 14/26 & 103 & 23.91 \\
% \quad Claude Fable 5.1  & \textbf{31.4} & \textbf{50.8} & \textbf{45.2} & 14/26 & 103 & 23.91 \\
\quad Claude Fable 5.1 & \textbf{29.8} & \textbf{47.4} & \textbf{42.4} & \underline{15/26} & 104 & 23.96 \\
% \quad Claude Opus 5     & \underline{29.0} & 45.2 & \underline{40.5} & \underline{13/26} & 166 & 17.67 \\
% \quad Claude Opus 5     & 28.5 & 45.2 & \underline{37.1} & \underline{13/26} & 166 & 17.67 \\
\quad Claude Opus 5 & 27.3 & \underline{42.9} & \underline{36.2} & \underline{15/26} & 166 & 17.56 \\
% \quad Claude Opus 4.8   & 19.8 & 45.5 & 34.3 & 15/26 & 91 & 9.46 \\
% \quad Claude Opus 4.8   & 18.8 & 45.5 & 29.2 & 15/26 & 91 & 9.46 \\
\quad Claude Opus 4.8 & 18.4 & 42.7 & 26.9 & 16/26 & 93 & 9.75 \\
\midrule
\rowcolor{modelgray}
\multicolumn{7}{l}{\textit{Codex}} \\
% \quad GPT-6-Astra       & 28.4 & \underline{50.5} & 30.9 & \textbf{10/26} & 147 & 24.18 \\
% \quad GPT-6-Astra       & \underline{28.8} & \underline{50.5} & 29.7 & \textbf{10/26} & 147 & 24.18 \\
\quad GPT 6 Astra & \underline{28.7} & \textbf{47.4} & 29.4 & \textbf{10/26} & 149 & 24.62 \\
% \quad GPT-5.6 Sol       & 21.8 & 44.8 & 29.6 & 15/26 & 57 & 2.26 \\
% \quad GPT-5.6 Sol       & 20.9 & 44.8 & 26.4 & 15/26 & 57 & 2.26 \\
\quad GPT 5.6 Sol & 19.4 & 41.9 & 25.9 & 17/26 & 55 & 2.23 \\
\midrule
\rowcolor{modelgray}
\multicolumn{7}{l}{\textit{OpenHands}} \\
% \quad DeepSeek V4 Flash & 17.8 & 31.2 & 32.0 & 14/26 & 215 & 1.25 \\
% \quad DeepSeek V4 Flash & 16.3 & 31.2 & 27.4 & 14/26 & 215 & 1.25 \\
\quad DeepSeek V4 Flash & 14.6 & 29.9 & 24.4 & \underline{15/26} & 206 & 1.24 \\
% \quad Kimi K3           & 14.6 & 25.3 & 30.6 & 15/26 & 196 & 7.70 \\
% \quad Kimi K3           & 13.1 & 25.3 & 27.7 & 15/26 & 196 & 7.70 \\
\quad Kimi K3 & 12.9 & 28.8 & 25.2 & 16/26 & 185 & 7.19 \\
\midrule
Human scientist        & 63.7 & 48.8 & 69.7 & 0/26 & -- & -- \\
\bottomrule
\end{tabular}
\end{adjustbox}
\caption{
\textbf{Main results on \ourbenchmark.}
PA and SI are reported unconditionally; SC failures count completed problems that fail either judge's scientific-constraint tests, out of 26 problems.
% \jy{Current results cover 23/26 problems, with full results expected in about two days; should we remove runtime or include post-evaluation time, which usually takes several hours?}
% \zw{add a subtitle above as "scientific value" and subtitle these as "efficiency"? "do we want to report the "SC pass rate" too?}
}
\label{tab:main_results}
\vspace{-3mm}
\end{table}

% \subsection{Accuracy Overstates Scientific Capability}
\label{sec:opt-metrics}

% \zw{looks like the main finding is agents do poorly on scientific discovery?}

\paragraph{AI agents still struggle to find the underlying mechanism.}
% In Table~\ref{tab:main_results}, our results show that AI agents struggle to find mechanisms that explain the initial observations and can be interpreted to derive novel scientific insights. We observe that the conditioned scores range from 14.6\% to 31.4\% across the seven models. Even the best-performing model, Claude Fable 5.1, achieves only 31.4\%, compared with 63.7\% for human-discovered mechanisms.
% In Table~\ref{tab:main_results}, our results show that AI agents struggle to find mechanisms that explain the initial observations and can be interpreted to derive novel scientific insights \bk{all the way into the result section, I am still not clear what mechanism means, I got some insights about "insights" in section 2. With only mention of insights for one field, and no mention of the 26 problems so far (maybe little bit in Figure 2), the manuscript becomes a bit murky on what exactly is studied/addressed here}. We observe that the conditioned scores range from 13.1\% to 31.4\% across the seven models. Even the best-performing model, Claude Fable 5.1, achieves only 31.4\%, compared with 63.7\% for human-discovered mechanisms.
The results in Table~\ref{tab:main_results} show that AI agents struggle to find mechanisms that explain the initial observations.
% \bk{all the way into the result section, I am still not clear what mechanism means, I got some insights about "insights" in section 2. With only mention of insights for one field, and no mention of the 26 problems so far (maybe little bit in Figure 2), the manuscript becomes a bit murky on what exactly is studied/addressed here}. 
The conditioned scores range from 12.9\% to 29.8\% across the seven AI agents. Even the best-performing agent, Claude Fable 5.1, achieves only 29.8\%, compared with 63.7\% for human-discovered mechanisms.
% Claude Opus 5 and GPT-6-Astra score 29.0\% and 28.4\%, respectively, while the remaining four models score at most 21.8\%.
% GPT-6-Astra and Claude Opus 5 score 28.8\% and 28.5\%, respectively, while the remaining four models score at most 20.9\%.
% The GPT-6-Astra and Claude Opus 5 have the similar score with Claude Fable 5.1 29.5\% and 28.4\%, respectively, while the remaining four models score at most 20.2\%.
% Fable fails at least one judge's gate on 14 problems, lowering its overall score while PA and insight scores are reported without gating.
Surprisingly, except GPT 6 Astra (10/26), all agents fail to pass the scientific-constraint tests on more than half of the problems in \ourbenchmark, showing that even top AI agents cannot reliably discover mechanisms that satisfy the most basic common-sense assumptions.
% Together, we show that AI agents still struggle to discover scientifically valid mechanisms, even after long-horizon experimentation.
Our empirical results show that current AI agents still lack the ability to reliably discover mechanisms that explain observations through the scientific experiment loop, which is a core capability for scientific discovery.

\paragraph{Agentic ability to discover scientific insights.}
To study the causes of this capability gap, we report predictive accuracy and SI scores separately in Table~\ref{tab:main_results}. We find that AI agents intentionally treat scientific mechanism discovery as an optimization task. Claude Fable 5.1 and GPT 6 Astra both achieve a predictive accuracy of 47.4\%, compared with 48.8\% for human solutions. However, their SI scores are only 42.4\% and 29.4\%, compared with 69.7\% for human scientists. This large gap in insight discovery demonstrates that human scientists value the interpretation and extrapolation of mechanisms, while AI agents tend to prioritize brute-force numerical optimization. Therefore, optimizing predictive accuracy does not imply discovering novel scientific insights (Appendix~\ref{app:pa_insights_stats}).

\section{Analysis}
\label{sec:analysis}
% \input{tables_dev/ablation}
% \zw{can we summarize the 3 analysis into the sec title? e.g., 'Analysis: What Bottlenecks Scientific Insight Discovery?', and subtitles "providing additional observation', ...}

\begin{figure}[h]
    \vspace{-10pt}
    \centering
    \small 
    \includegraphics[width=0.85\linewidth]{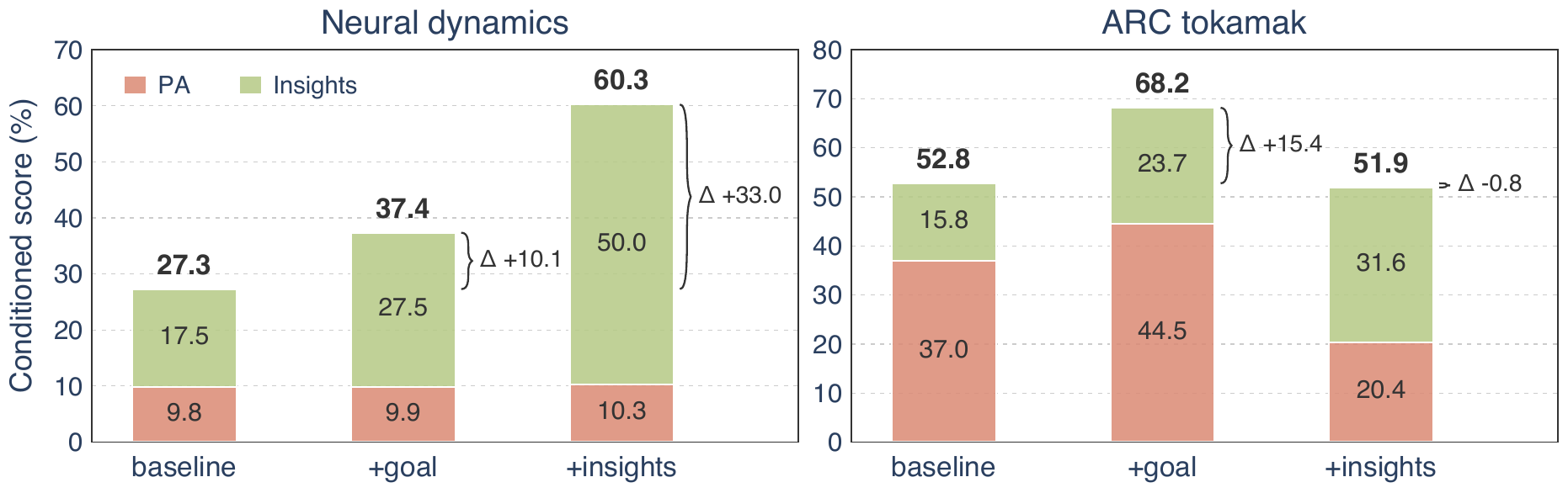}
    \captionsetup{font=small,skip=6pt}
    \caption{
       \textbf{Effects of goals and insight questions on mechanism discovery.}
       We evaluate Claude Fable 5.1 with general guidance (+goal) or insight questions (+insights); stacked bars show estimated PA and SI contributions to the conditioned score, with $\Delta$ marking changes from baseline.
       % \zw{the stacked bars are a bit hard to understand, why separate bars cannot convey our findings?}
       }
    \label{fig:context_ablation}
\end{figure}

\paragraph{Does additional guidance improve mechanism discovery?} 
% \zw{since we only have this main finding in this subsec, why not directly move it to the title?}
% We study whether additional guidance helps AI agents discover mechanisms through the long-horizon experimentation loop.
We study two types of guidance: enabling the ``goal'' feature (+goal) and directly including insight questions in the task specification (+insights).
We evaluate Claude Fable 5.1, the best-performing agent, on one neuroscience problem (\textit{neural dynamics}) and one plasma physics problem (\textit{ARC tokamak}), as shown in Figure~\ref{fig:context_ablation}.
In the +goal condition, we specify that a good mechanism should be not only accurate but also useful for future exploration (Appendix~\ref{app:goal_prompt}).
Enabling this feature increases the SI contribution to the conditioned score on \textit{neural dynamics} ($\Delta +10.0\%$) and \textit{ARC tokamak} ($\Delta +7.9\%$).
Directly providing insight questions in the task specification as explicit optimization targets substantially increases the SI contribution on \textit{neural dynamics} ($\Delta +32.5\%$) and \textit{ARC tokamak} ($\Delta +15.8\%$).
However, this is not a natural setting: when scientists first observe a phenomenon that challenges their understanding, they do not yet know what insights its underlying mechanism will yield.

% On \textit{arc tokamak}\bk{not defined, also both ARC and \textit{arc} are used} the conditioned score improves only slightly ($\Delta +4.9$), and the gain comes from predictive accuracy rather than insights.
% \subsection{Mechanism-Derived and Agent-Discovered Insights}
% \label{sec:insights_during_discovery}
% End the wrap immediately below the caption so subsequent lines use the full width.
\begin{wrapfigure}[15]{r}{0.6\columnwidth}
    \centering
    \includegraphics[width=0.95\linewidth]{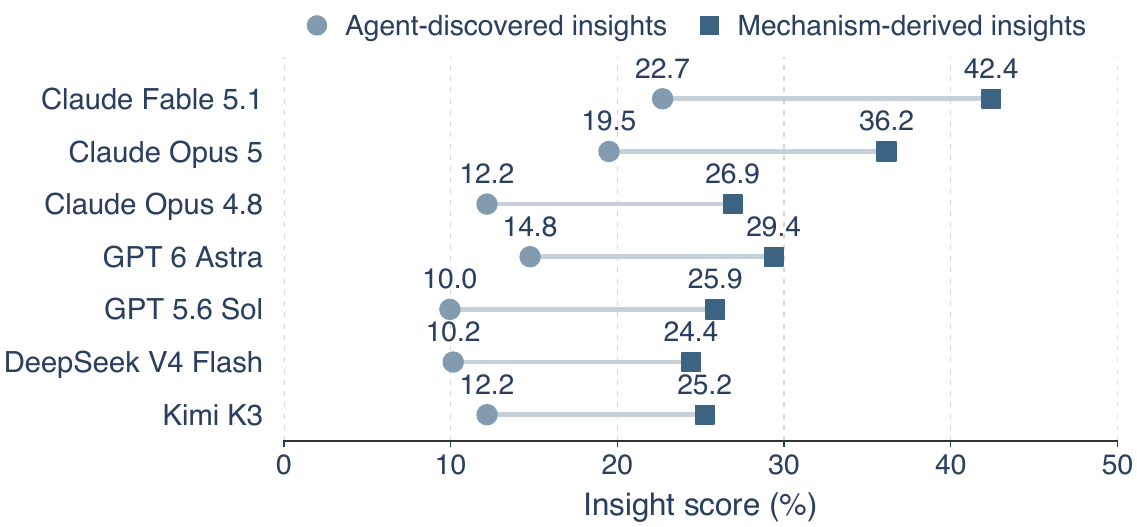}
    \captionsetup{skip=2pt}
    \caption{
        \textbf{Insights discovered during the experimentation loop vs. those derived from the discovered mechanism.}
        % We compare insight scores based on agents' findings during experimentation with scores obtained by interpreting and extrapolating their submitted mechanisms.
        Both scores use the same problems and instances.
    }
    \label{fig:insight_discovery}
\end{wrapfigure}

\paragraph{How many insights are derived from the discovered mechanism?}
% We study how many insights agents derive during their experiments by asking the judge agent to evluate their submitted mechanism without simulator access (Figure~\ref{fig:insight_discovery}).
To assess insights agents derive during experimentation, the judge evaluates their submitted mechanisms without simulator access (Figure~\ref{fig:insight_discovery}).
All agents score lower than in the main evaluation, where the judge interprets and extrapolates their mechanisms using the simulator.
% Claude Fable 5.1 has the highest score for findings from experimentation at 21.0\%, compared with 40.4\% from its mechanism.
Claude Fable 5.1 scores highest and shows the largest gap: 22.7\% from experimentation versus 42.4\% from its mechanism.
% Claude Opus 4.8 shows the largest difference, scoring 10.4\% on its own findings and 31.4\% when its mechanism is explored further.
We also observe that, in 6.7\% of cases, judges credit insights from experiment logs that they cannot derive from the mechanism.
This comparison lets us evaluate insights derived by extrapolating the mechanism separately from insights agents find during experiments.

\begingroup
\setlength{\intextsep}{12pt}
\begin{figure}[htbp]
    \centering
    \small
    \includegraphics[width=\textwidth]{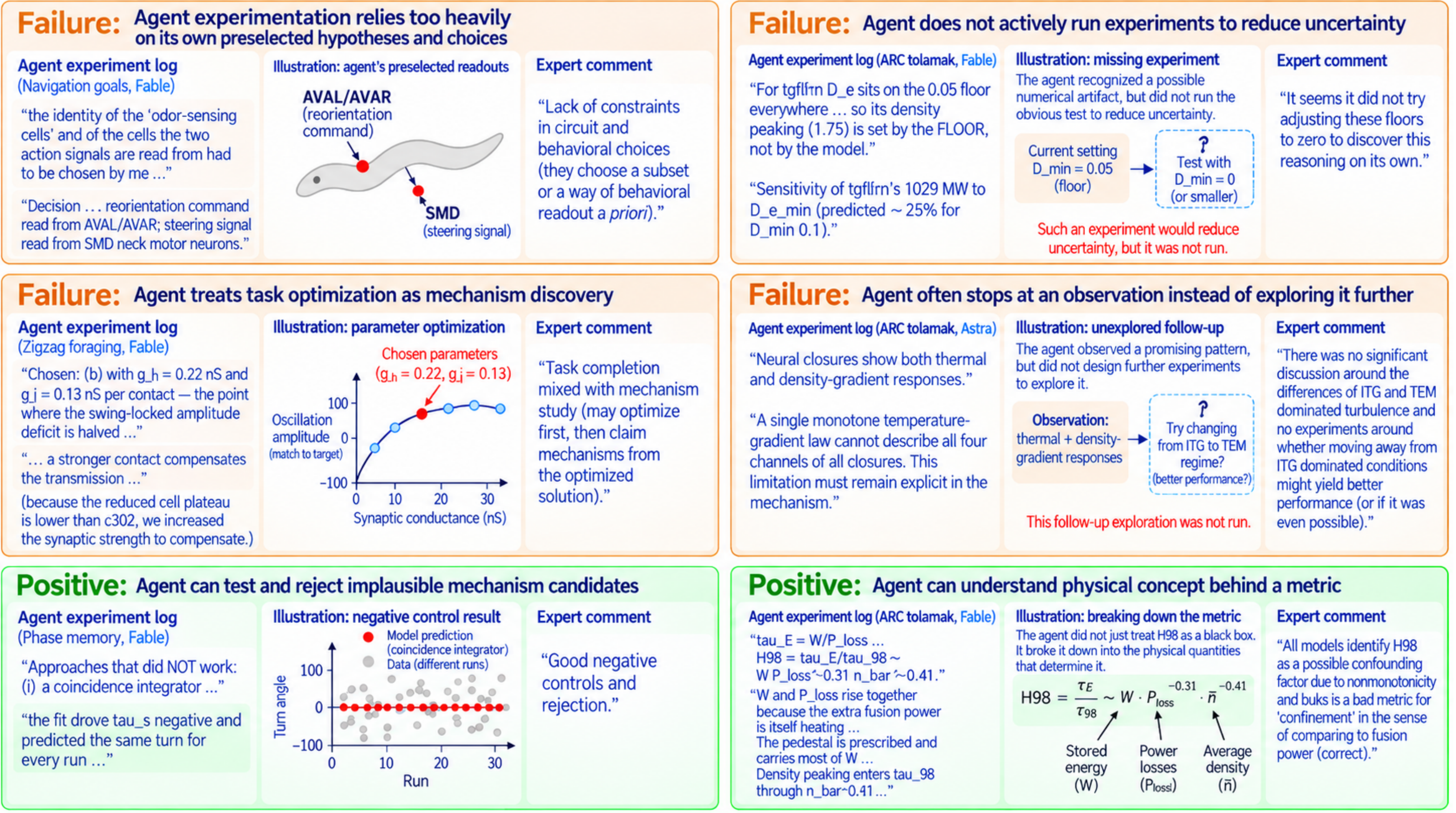}
    \captionsetup{skip=8pt}
    \caption{
        \textbf{Expert review of mechanism discovery.}
        Experts review experiment logs of Claude Fable 5.1 and GPT 6 Astra on neuroscience and plasma physics tasks, highlighting failures in mechanism discovery alongside successful hypothesis rejection and understanding of physical concepts.
    }
    \label{fig:failure_analysis}
\end{figure}
\endgroup

\paragraph{Qualitative analysis: where do agents' experiments fall short?}
\phantomsection\label{par:qualitative_analysis}
Our experts review agents' mechanisms and experiment logs and identify four types of failure that hinder the discovery of novel scientific insights (Figure~\ref{fig:failure_analysis}).
\textbf{\textit{(1) Overreliance on preselected hypotheses and choices.}}
Agents build explanations around unjustified choices. In \textit{navigation goals}, the agent selects neuronal readouts without establishing whether they capture the relevant circuit and behavior.
\textbf{\textit{(2) Treating task optimization as mechanism discovery.}}
Agents equate fitting observations with explaining them. In \textit{zigzag foraging}, the agent adjusts synaptic strength to match a target response and presents the optimized result as a mechanism.
\textbf{\textit{(3) Not using experiments to resolve uncertainty.}}
Claude Fable 5.1 recognizes a possible numerical artifact in \textit{ARC tokamak} but does not lower the diffusivity floor to test its effect.
\textbf{\textit{(4) Stopping at observations without further exploration.}}
GPT 6 Astra observes responses to temperature and density gradients in \textit{ARC tokamak} but does not explore whether changing the dominant turbulence regime could improve performance.
Nevertheless, agents sometimes make sound experimental decisions, such as rejecting mechanisms that contradict the evidence, and recognize the physical meaning of variables and model parameters.
These findings suggest that agents are better at task optimization than at open-ended experimentation to resolve scientific uncertainty.
We provide detailed and complete descriptions of the common failure modes in Appendix~\ref{app:common_failures}.
Common positive behaviors are described in Appendix~\ref{app:common_positives}.

\section{Related Work}
\label{sec:related_work}

\paragraph{AI scientist agents.}
% \zw{use paragraphs instead of subsec to save space?}
AI scientist agents automate research workflows, from reviewing literature and generating ideas to running experiments and writing papers \citep{lu2024ai,yamada2025ai,schmidgall2025laboratory,tang2025researcher}. AI agents also review scientific papers \citep{kim2026limits}. Recent systems incorporate persistent research memory and connect claims to evidence \citep{lyu2026evoscientist,meng2026scientistone}, while other work improves research capabilities through execution feedback and replication training \citep{si2026towards,falck2026training}. Applications beyond machine learning include laboratory experimentation and hypothesis validation \citep{boiko2023autonomous,gottweis2026accelerating,ghareeb2026robin}, as well as research that combines literature analysis with large-scale data exploration or multimodal observations \citep{mitchener2025kosmos,li2026omniscientist}. Related systems search for executable algorithms or equations using automated evaluators \citep{romeraparedes2024mathematical,novikov2025alphaevolve,shojaee2025llm}. For mechanism discovery, agents use experiments to investigate physical and computational systems and revise mechanistic models \citep{kabra2026llm,murphy2026model,wang2026codeworlds,wang2026mechanist}.

\paragraph{Benchmarking scientific discovery.}
Existing benchmarks evaluate scientific tool use and research workflows \citep{chen2024scienceagentbench,bragg2026astabench,liu2026sciagentarena}, verifiable outputs and scientific software engineering \citep{terminalbenchscience2026,xu2026swe}, experiment replication \citep{starace2025paperbench,kon2025expbench,qiu2026prbench}, or gains over numerical baselines \citep{chan2024mlebench,nathani2025mlgym,lupidi2026airs,wang2026naturebench}. DiscoveryBench and LLM-SRBench assess hypotheses and equations recovered from data \citep{majumder2024discoverybench,shojaee2025llmsrbench}, while interactive benchmarks let agents gather observations to uncover hidden systems and physical laws \citep{geng2025large,koblischke2025gravitybench,chen2025physgym,zheng2025newtonbench,jain2026madphysics}. DiscoveryWorld evaluates explanatory knowledge, and BoxingGym tests whether explanations help another agent predict unseen outcomes \citep{jansen2024discoveryworld,gandhi2025boxinggym}. Closest to our setting, DiscoverPhysics evaluates both trajectory prediction and conceptual explanations against expert-written rubrics in simulated worlds with noncanonical physics \citep{wiemann2026discoverphysics}. Unlike benchmarks centered on task completion or numerical performance, we focus on a core capability of scientific discovery: whether agents can uncover scientifically meaningful insights while seeking an answer, rather than merely produce a correct final answer.

\section{Conclusion}
\label{sec:discussion}

As \ourbenchmark demonstrates, the answer to the question of whether AI agents can understand mechanisms and uncover novel scientific insights is still unclear. While agents excel at optimizing predictive accuracy, on many problems their mechanisms fail to satisfy even basic common-sense assumptions, and they often rely on untested assumptions, mistake task optimization for mechanism discovery, leave uncertainty unresolved, and stop without further exploration. Given these findings, we believe that \ourbenchmark opens up a new path for the scientific study of AI-driven mechanism discovery across a variety of scientific fields, and leaves open a number of new challenges:

\paragraph{Multiple paths in scientific discovery.}
Scientific discovery can follow different paths from the same initial observations, leading to different mechanisms and scientific insights. Our insight questions reflect the research directions selected with domain experts and capture only part of what can be learned about each problem. For example, an agent might find that its mechanism explains another phenomenon or predicts an effect we had not considered when designing the benchmark. These findings would not count toward its SI score unless they answer one of the questions in our rubric. Future work could expand the insight set to cover alternative paths, with domain experts assessing findings that extend beyond established knowledge.

\paragraph{Evaluating scientific insights.}
Our SI evaluation requires judge agents to interpret and extrapolate mechanisms, which means it is also constrained by their own scientific reasoning abilities. Judges may accept confidently stated conclusions without adequately establishing how they follow from the mechanism, potentially overestimating the number of scientific insights it supports. The expert reviews in our \hyperref[par:qualitative_analysis]{qualitative analysis} suggest that some automated assessments are overly generous, although these reviews provide broad judgments of selected submissions rather than scores for every insight question. On \textit{ARC tokamak}, our fusion expert rated Claude Fable 5.1's submission above the other submissions they had reviewed, but considered its work closer to that of a master's student in the domain than to that of a researcher.

\paragraph{Training agents to discover scientific insights.}
One way to improve agents' ability to discover new scientific insights is to build reinforcement learning (RL) environments that train them to do so through experimentation. Rewards could reflect whether agents test their proposed mechanisms and support their scientific insights with experimental evidence. Such training may help agents choose informative experiments and investigate what their mechanisms imply beyond the initial observations.

\section{Acknowledgments}
This work was supported in part by grants from the US Department of Energy and Fujitsu. We thank Sean Welleck for helpful early discussions, and Amanda Bertsch, Lintang Sutawika, Xueqiao Sun, Zedian (Mark) Xiao, and Yixiong Fang for their valuable feedback on the project. We also thank Ruyi Ke from Princeton University's Department of Chemistry, Yuyang Wu from Carnegie Mellon University's Department of Chemistry, Yiming Yang from UC Berkeley's Department of Physics, and Drew Bridges and Seh Na Mellick from Carnegie Mellon University's Department of Biological Sciences. We thank Christine Kaeser-Chen and the fusion team at Google DeepMind.

\newpage
\bibliographystyle{iclr2027_conference}
% \bibliography{ref}
\bibliography{iclr2027_conference}

\newpage

% Lite appendix: retain instructions, results, problem backgrounds, and environment interfaces.
\appendix

% \section{Task Specification Example}
\Needspace{40\baselineskip}
\section{Task Specification Template}
\label{app:task_specification}
% Source: p-agent-mechanism-discovery/prompts/neuroscience/elegans.yaml, neural_dynamics.
% Instruction wording is copied from the source; placeholders replace problem-specific content.
\begingroup
\definecolor{promptaccent}{HTML}{3C6382}
\fontsize{9.5}{11.5}\selectfont
\setlength{\tabcolsep}{8pt}
\setlength{\LTleft}{0pt}
\setlength{\LTright}{0pt}
\setlength{\LTcapwidth}{\textwidth}
\setlength{\LTpre}{0pt}
\setlength{\LTpost}{0pt}
\renewcommand{\arraystretch}{1.1}
\arrayrulecolor{promptaccent}
\begin{longtable}{@{}>{\raggedright\arraybackslash}p{0.17\textwidth}>{\raggedright\arraybackslash}p{\dimexpr0.83\textwidth-2\tabcolsep\relax}@{}}
\toprule[0.7pt]
\rowcolor{promptaccent!8}
\multicolumn{2}{l}{\textcolor{promptaccent}{\textbf{Task specification}}} \\[3pt]
\endfirsthead
\toprule[0.7pt]
\rowcolor{promptaccent!8}
\multicolumn{2}{l}{\textcolor{promptaccent}{\textbf{Task specification} (continued)}} \\[3pt]
\endhead
\arrayrulecolor{promptaccent}\bottomrule[0.7pt]
\endfoot
\arrayrulecolor{promptaccent}\bottomrule[0.7pt]
\caption{\textbf{Task specification template for the scientist agent (neuroscience).}}\label{tab:task_specification_template} \\
\endlastfoot
\arrayrulecolor{promptaccent!25}\midrule[0.3pt]
\textcolor{promptaccent}{\textbf{Context}} & You are a systems neuroscientist studying \textcolor{promptaccent}{\texttt{\{problem\_description\}}}. \\[5pt]
 & \textcolor{promptaccent}{\texttt{\{scientific\_background\}}} \textcolor{promptaccent}{\texttt{\{observations\_block\}}} \\[5pt]
\arrayrulecolor{promptaccent!25}\midrule[0.3pt]
\textcolor{promptaccent}{\textbf{Initial observations}} & However, it is unclear \textcolor{promptaccent}{\texttt{\{initial\_observations\}}}. \\[5pt]
\arrayrulecolor{promptaccent!25}\midrule[0.3pt]
\textcolor{promptaccent}{\textbf{Task}} & Your task is to discover the mechanism that explains this phenomenon. \textcolor{promptaccent}{\texttt{\{mechanism\_requirements\}}} To discover this, you interrogate \textcolor{promptaccent}{\texttt{\{environment\_interface\}}} through controlled numerical experiments designed to identify \textcolor{promptaccent}{\texttt{\{experimental\_objectives\}}}. The structure of your predictive form needs to follow from the mechanism you uncover: for each ingredient, explain why it is included, what it traces in the circuit, and how it shapes the neuron's activity. \\[5pt]
\arrayrulecolor{promptaccent!25}\midrule[0.3pt]
\textcolor{promptaccent}{\textbf{Evaluation}} & Every claim in your discovered mechanism needs to be backed by data you actually collected. \textcolor{promptaccent}{\texttt{\{predictive\_accuracy\_evaluation\}}} Your experimental findings will also be reviewed by an independent researcher who has access to the same simulator, to assess whether the mechanism you discover can yield new insights. \\[5pt]
\arrayrulecolor{promptaccent!25}\midrule[0.3pt]
\textcolor{promptaccent}{\textbf{Deliverables}} & You need to produce three deliverables. If any of them is missing or does not follow the requirements below, your solution receives a score of 0.\newline Deliverable 1:\newline Write your discovered mechanism to \textcolor{promptaccent}{\texttt{\{output\_dir\}}}/mechanisms/mechanism.md. State the mechanism as the causal chain from the coarse action state of the population to the activity of one neuron; its executable form, with every ingredient and fitted constant defined and the physical process each one traces identified; why its inputs jointly determine the neuron's activity; the concrete evidence from your experiments that supports the mechanism and its form, including the evidence that let you reject the competing explanations you considered; and the insights about how population activity organizes the activity of individual neurons that follow from what you found. \\[5pt]
 & Deliverable 2:\newline Write the executable form of your mechanism to \textcolor{promptaccent}{\texttt{\{output\_dir\}}}/mechanisms/mechanism.py.\newline \textcolor{promptaccent}{\texttt{\{executable\_interface\_and\_requirements\}}} \\[5pt]
 & Deliverable 3:\newline Keep an experiment log at \texttt{\{output\_dir\}/mechanisms/experiment.log}. Treat it as your laboratory notebook: it is the primary reference for the researchers who will continue this work, and you choose how to organize it. Record what they will need to follow and extend your investigation: the questions you pursued, the simulator configurations and analyses you ran, what you observed, the conclusions you drew, and the approaches that did not work and why. Give particular attention to the findings that surprised you or that you found interesting along the way, whether or not they entered your final mechanism; these are often the most valuable part of a notebook for the next researcher. Move the useful evidence behind the log's findings --- the figures, analysis outputs, and small data files your experiments produced in \textcolor{promptaccent}{\texttt{\{tmp\_dir\}}}, and the observations from \texttt{\{output\_dir\}/observations/} that your final claims build on --- into \texttt{\{output\_dir\}/mechanisms/} so that they survive for these researchers: everything outside \texttt{\{output\_dir\}/mechanisms/} is deleted when the session ends. \\[5pt]
\arrayrulecolor{promptaccent!25}\midrule[0.3pt]
\textcolor{promptaccent}{\textbf{Saved evidence}} & You may additionally place supporting evidence in \texttt{\{output\_dir\}/mechanisms/} and reference it from the files above: figures, and any analysis outputs that back your claims, such as fitted-model diagnostics, ablation results, or small data tables worth preserving. Only this folder is kept when the session ends; everything else in the workspace, including \texttt{\{output\_dir\}/observations/}, \textcolor{promptaccent}{\texttt{\{tmp\_dir\}}} and the simulation outputs you generated, is discarded. The independent researcher reviewing your mechanism sees exactly this folder. Everything under \texttt{\{output\_dir\}/mechanisms/} combined is limited to 1 GB: if the folder exceeds this limit, your solution receives a score of 0. It is for final deliverables and their evidence, not scratch space; keep drafts and intermediate files in \textcolor{promptaccent}{\texttt{\{tmp\_dir\}}}. \\[5pt]
\arrayrulecolor{promptaccent!25}\midrule[0.3pt]
\textcolor{promptaccent}{\textbf{Budget}} & You have a time budget of \textcolor{promptaccent}{\texttt{\{time\_budget\}}}. You can use \textcolor{promptaccent}{\texttt{\{tmp\_dir\}}} for notes and intermediate files. \\[5pt]
\end{longtable}
\endgroup

\Needspace{26\baselineskip}
\section{Guidance Analysis Prompts}
\label{sec:appendix}
\label{app:analysis_settings}
\subsection{Goal Prompt}
\label{app:goal_prompt}
\begin{table}[h]
    \centering
    \definecolor{goalaccent}{HTML}{3C6382}
    \setlength{\tabcolsep}{8pt}
    \setlength{\extrarowheight}{2pt}
    \renewcommand{\arraystretch}{1.12}
    \arrayrulecolor{goalaccent}
    \begin{tabularx}{\textwidth}{>{\raggedright\arraybackslash}p{0.20\textwidth}>{\raggedright\arraybackslash}X}
        \toprule[0.65pt]
        \rowcolor{goalaccent!8}
        \textcolor{goalaccent}{\textbf{Problem}} & \textcolor{goalaccent}{\textbf{Goal}} \\[2pt]
        \arrayrulecolor{goalaccent!35}\midrule[0.3pt]
        % Source: p-agent-mechanism-discovery/scripts/neuroscience/neural_dynamics/run_all_goal.sh (first two sentences of goal_text).
        \textcolor{goalaccent}{\textbf{Neural dynamics}} & A good mechanism is not only accurate but also extrapolatable, and it can even open up more interesting new research directions. Through such a mechanism people come to understand how the nervous system of C. elegans organizes its activity into a hierarchical locomotion structure, down to how the coarse action state shapes a single neuron. \\[4pt]
        \midrule[0.3pt]
        % Source: p-agent-mechanism-discovery/scripts/fusion_plasma/arc_tokamak/run_all_goal.sh (first two sentences of goal_text).
        \textcolor{goalaccent}{\textbf{ARC tokamak}} & A good mechanism is not only accurate but also extrapolatable, and it can even open up more interesting new research directions. Through such a mechanism people come to understand how the transport a model returns sets the profiles an ARC plasma settles into and the fusion power it produces, down to why two models that leave the plasma with the same energy confinement disagree by a factor of two in power. \\[4pt]
        \arrayrulecolor{goalaccent}\bottomrule[0.65pt]
    \end{tabularx}
    \caption{\textbf{Goal guidance used in the \texttt{+goal} condition.}}
    \label{tab:goal_prompts}
\end{table}

\Needspace{48\baselineskip}
\section{Agentic Judge Prompts}
\label{app:evaluation}
\subsection{Scientific Constraint Evaluation}
\label{app:sc_judge_prompt}
% Source: p-agent-mechanism-discovery/prompts/neuroscience/elegans_judge.yaml, neural_dynamics_sc_c.
% Instruction wording is copied from the source; placeholders replace problem-specific content.
\noindent\begin{minipage}{\textwidth}
\centering
\definecolor{promptaccent}{HTML}{3C6382}
\fontsize{9.5}{11.5}\selectfont
\setlength{\tabcolsep}{8pt}
\renewcommand{\arraystretch}{1.1}
\arrayrulecolor{promptaccent}
\begin{tabularx}{\textwidth}{@{}>{\raggedright\arraybackslash}p{0.17\textwidth}>{\raggedright\arraybackslash}X@{}}
\toprule[0.7pt]
\rowcolor{promptaccent!8}
\multicolumn{2}{l}{\textcolor{promptaccent}{\textbf{Scientific constraint judge instructions}}} \\[3pt]
\arrayrulecolor{promptaccent!25}\midrule[0.3pt]
\textcolor{promptaccent}{\textbf{Context}} & You are a systems neuroscientist reviewing another neuroscientist's discovered mechanism for \textcolor{promptaccent}{\texttt{\{problem\_description\}}}. \\[5pt]
 & The neuroscientist interrogated \textcolor{promptaccent}{\texttt{\{environment\_interface\}}} through controlled simulations to discover that mechanism. \\[5pt]
\arrayrulecolor{promptaccent!25}\midrule[0.3pt]
\textcolor{promptaccent}{\textbf{Verdict instruction}} & Your task is to judge whether this mechanism is consistent with commonly accepted assumptions about C. elegans. The criteria you need to check are listed below: \\[5pt]
 & Commonly accepted assumptions (Scientific Constraints)\newline According to the discovered mechanism:\newline \textcolor{promptaccent}{\texttt{\{scientific\_constraints\_criteria\}}} \\[5pt]
 & Each of these criteria is a general assumption, not something the neuroscientist is expected to prove. Judge each one by reading the mechanism described in mechanism.md together with its executable form. A criterion should be marked as false if the mechanism contradicts the assumption, and true if it does not. Focus on what the mechanism actually does, not on the names given to its quantities; calling a quantity something is not enough to show that it behaves that way. You also need to mark the criterion as false if the experimental results violate the assumption and the neuroscientist neither recognizes those results as erroneous nor accounts for the violation in the final mechanism. \\[5pt]
\arrayrulecolor{promptaccent!25}\midrule[0.3pt]
\textcolor{promptaccent}{\textbf{Output}} & You are expected to provide the following:\newline Write \textcolor{promptaccent}{\texttt{\{output\_dir\}}}/\textcolor{promptaccent}{\texttt{\{judge\_output\}}} as a JSON file with exactly the structure below, with one entry for each criterion ID listed above and nothing else: \\[5pt]
 & \texttt{\{}\newline \hspace*{2em}\texttt{"scientific\_constraints": \{}\newline \hspace*{4em}\texttt{"SC1": \{"verdict": true, "reason": "..."\}}\newline \hspace*{2em}\texttt{\}}\newline \texttt{\}} \\[5pt]
 & Each verdict should be either true or false, with no other values allowed. Each reason should be no more than two sentences and should identify the part of the submission that supports the verdict. Keep \textcolor{promptaccent}{\texttt{\{output\_dir\}}}/judge\_notes.md alongside the JSON file and use it to record what you reviewed for each criterion. Save both files instead of printing their contents, then stop. \\[5pt]
\arrayrulecolor{promptaccent}\bottomrule[0.7pt]
\end{tabularx}
\captionof{table}{\textbf{Example instructions for scientific constraint tests (neuroscience).}}
\label{tab:sc_judge_prompt}
\end{minipage}

\Needspace{52\baselineskip}
\subsection{Judge Instructions for Insights Derived from the Mechanism}
\label{app:insight_judge_prompt}
% Source: p-agent-mechanism-discovery/prompts/neuroscience/elegans_judge.yaml, neural_dynamics_insights_strict.
% Instruction wording is copied from the source; placeholders replace problem-specific content.
\noindent\begin{minipage}{\textwidth}
\centering
\definecolor{promptaccent}{HTML}{3C6382}
\fontsize{9.5}{11.5}\selectfont
\setlength{\tabcolsep}{8pt}
\renewcommand{\arraystretch}{1.1}
\arrayrulecolor{promptaccent}
\begin{tabularx}{\textwidth}{@{}>{\raggedright\arraybackslash}p{0.17\textwidth}>{\raggedright\arraybackslash}X@{}}
\toprule[0.7pt]
\rowcolor{promptaccent!8}
\multicolumn{2}{l}{\textcolor{promptaccent}{\textbf{Judge instructions for insights derived from the mechanism}}} \\[3pt]
\arrayrulecolor{promptaccent!25}\midrule[0.3pt]
\textcolor{promptaccent}{\textbf{Context}} & You are a systems neuroscientist reviewing another neuroscientist's discovered mechanism for \textcolor{promptaccent}{\texttt{\{problem\_description\}}}. \\[5pt]
 & The neuroscientist interrogated \textcolor{promptaccent}{\texttt{\{environment\_interface\}}} through controlled simulations to discover that mechanism. \\[5pt]
\arrayrulecolor{promptaccent!25}\midrule[0.3pt]
\textcolor{promptaccent}{\textbf{Verdict instruction}} & A good mechanism is not only accurate but also far more interpretable and extrapolatable, and it can be used to derive insights. Your task is strictly to evaluate whether this discovered mechanism is able to derive the interpretable scientific insights listed below. \\[5pt]
 & Scientific insight questions\newline \textcolor{promptaccent}{\texttt{\{insights\_criteria\}}} \\[5pt]
 & To derive each insight, you need to start from the discovered mechanism and use \textcolor{promptaccent}{\texttt{\{environment\_interface\}}} to check whether the insight can be derived from it. A Yes verdict only occurs when the discovered mechanism is able to help you find the evidence needed to provide a concrete and complete explanation of each insight question. A Yes verdict requires the step-by-step chain by which the discovered mechanism's own components produce the insight in the nervous system itself; a result that holds only because of the stimuli chosen for the experiments, or that is only observed in the simulator or the saved data, is not such a derivation. You should give a No verdict for an insight question if the answer is based on your prior knowledge or is not fully supported by the discovered mechanism. \\[5pt]
 & Any assertion or conclusion about the insights in mechanism/ is never, by itself, grounds for a Yes verdict; verify that the mechanism derives the same conclusion. If you are not able to provide sufficient evidence showing that this mechanism is able to derive the insight, the verdict should be No. \\[5pt]
 & You must not modify or extend the discovered mechanism. If the insight question cannot be completely derived from the discovered mechanism, the verdict should be No. An answer based only on your own knowledge of the animal or on the simulator, but not supported by the discovered mechanism, should also be No. \\[5pt]
\arrayrulecolor{promptaccent!25}\midrule[0.3pt]
\textcolor{promptaccent}{\textbf{Output}} & You are expected to provide the following:\newline Write \textcolor{promptaccent}{\texttt{\{output\_dir\}}}/\textcolor{promptaccent}{\texttt{\{judge\_output\}}} as a JSON file with exactly the structure below, with one entry for each criterion ID listed above and nothing else: \\[5pt]
 & \texttt{\{}\newline \hspace*{2em}\texttt{"insights": \{}\newline \hspace*{4em}\texttt{"I1": \{"verdict": "Yes", "reason": "..."\}}\newline \hspace*{2em}\texttt{\}}\newline \texttt{\}} \\[5pt]
 & Each verdict should be either Yes or No, with no other values allowed. For each insight criterion, when you give a Yes verdict, you must provide complete evidence and reasoning explaining why the discovered mechanism can derive the insight asked about in the insight question; otherwise, the verdict should be No. Keep \textcolor{promptaccent}{\texttt{\{output\_dir\}}}/judge\_notes.md alongside the JSON file and use it to record what you reviewed and any commands you ran. Save both files instead of printing their contents, then stop. \\[5pt]
\arrayrulecolor{promptaccent}\bottomrule[0.7pt]
\end{tabularx}
\captionof{table}{\textbf{Example judge instructions for insights derived from the mechanism (neuroscience).} The judge returns \texttt{Yes} only with complete evidence showing how the insight can be derived from the discovered mechanism.}
\label{tab:insight_judge_prompt}
\end{minipage}

\Needspace{52\baselineskip}
\subsection{Judge Instructions for Insights Discovered During Experimentation}
\label{app:judge_discovered_insights}
% Source: p-agent-mechanism-discovery/prompts/neuroscience/elegans_judge.yaml, neural_dynamics_insights_read.
% Instruction wording is copied from the source; placeholders replace problem-specific content.
\noindent\begin{minipage}{\textwidth}
\centering
\definecolor{promptaccent}{HTML}{3C6382}
\fontsize{9.5}{11.5}\selectfont
\setlength{\tabcolsep}{8pt}
\renewcommand{\arraystretch}{1.1}
\arrayrulecolor{promptaccent}
\begin{tabularx}{\textwidth}{@{}>{\raggedright\arraybackslash}p{0.17\textwidth}>{\raggedright\arraybackslash}X@{}}
\toprule[0.7pt]
\rowcolor{promptaccent!8}
\multicolumn{2}{l}{\textcolor{promptaccent}{\textbf{Judge instructions for insights discovered during experimentation}}} \\[3pt]
\arrayrulecolor{promptaccent!25}\midrule[0.3pt]
\textcolor{promptaccent}{\textbf{Context}} & You are a systems neuroscientist reviewing another neuroscientist's discovered mechanism for \textcolor{promptaccent}{\texttt{\{problem\_description\}}}. \\[5pt]
 & The neuroscientist interrogated \textcolor{promptaccent}{\texttt{\{environment\_interface\}}} through controlled simulations to discover that mechanism. \\[5pt]
\arrayrulecolor{promptaccent!25}\midrule[0.3pt]
\textcolor{promptaccent}{\textbf{Verdict instruction}} & Your task is to evaluate whether the scientific insight captured by each question below was actually discovered by the neuroscientist while uncovering the mechanism. Consider the questions one at a time. For each question, determine whether the neuroscientist obtained the corresponding insight and whether the available evidence is strong enough to support it. The scientific insight questions are listed below: \\[5pt]
 & Scientific insight questions\newline \textcolor{promptaccent}{\texttt{\{insights\_criteria\}}} \\[5pt]
 & When you judge each insight question, do not modify or extend the discovered mechanism, and do not look for a different one. The neuroscientist needs to have explicitly stated the insight and run an experiment specifically to test it. Experiments run for other purposes, such as building, tuning, or checking the executable form of the mechanism, do not count as tests even if their results bear on the question. For questions asking why something happens, showing that it happens is not enough; the explanation itself needs to have been tested. \\[5pt]
 & If the neuroscientist did not discover the insight while uncovering the mechanism, or if the discovered mechanism and experiment record do not provide sufficient evidence that the insight was obtained during this process, the answer should be No. If the insight needs to be inferred from their numbers or their argument needs to be completed for them, the answer should be No. An answer based only on your own knowledge of the animal, but not supported by a finding the neuroscientist recorded, should also be No. If the mechanism clearly conflicts with that insight, the answer should also be No. \\[5pt]
 & A multi-part question is answered only if every part is answered. Each underlying insight can earn only one Yes verdict. Answers based on the same explanation count as the same insight regardless of the quantities or experiments involved, so mark as Yes only the question that explanation answers most directly and the rest No. When in doubt, the answer should be No. \\[5pt]
\arrayrulecolor{promptaccent!25}\midrule[0.3pt]
\textcolor{promptaccent}{\textbf{Output}} & You are expected to provide the following:\newline Write \textcolor{promptaccent}{\texttt{\{output\_dir\}}}/\textcolor{promptaccent}{\texttt{\{judge\_output\}}} as a JSON file with exactly the structure below, with one entry for each criterion ID listed above and nothing else: \\[5pt]
 & \texttt{\{}\newline \hspace*{2em}\texttt{"insights": \{}\newline \hspace*{4em}\texttt{"I1": \{"verdict": "Yes", "reason": "..."\}}\newline \hspace*{2em}\texttt{\}}\newline \texttt{\}} \\[5pt]
% Original: Each verdict should be either true or false, with no other values allowed. Each reason should be no more than two sentences. A true verdict has to quote the neuroscientist's own statement of the insight and the measurement from the experiment that tested it, and has to say what the neuroscientist ran that experiment for, otherwise the verdict is false. Keep {output_dir}/judge_notes.md alongside the JSON file and use it to record what you reviewed for each criterion. Save both files instead of printing their contents, then stop.
 & Each verdict should be either Yes or No, with no other values allowed. Each reason should be no more than two sentences. A Yes verdict has to quote the neuroscientist's own statement of the insight and the measurement from the experiment that tested it, and has to say what the neuroscientist ran that experiment for; otherwise, the verdict is No. Keep \textcolor{promptaccent}{\texttt{\{output\_dir\}}}/judge\_notes.md alongside the JSON file and use it to record what you reviewed for each criterion. Save both files instead of printing their contents, then stop. \\[5pt]
\arrayrulecolor{promptaccent}\bottomrule[0.7pt]
\end{tabularx}
\captionof{table}{\textbf{Example judge instructions for insights discovered during experimentation (neuroscience).}}
\label{tab:judge_discovered_insights}
\end{minipage}

\Needspace{30\baselineskip}
\section{Additional Results}
\label{app:additional_results}
\subsection{How do predictive accuracy and scientific insights vary across domains?}
\label{app:insight_scores_across_domains}
\begin{figure}[htbp]
    \centering
    \small \includegraphics[width=\textwidth]{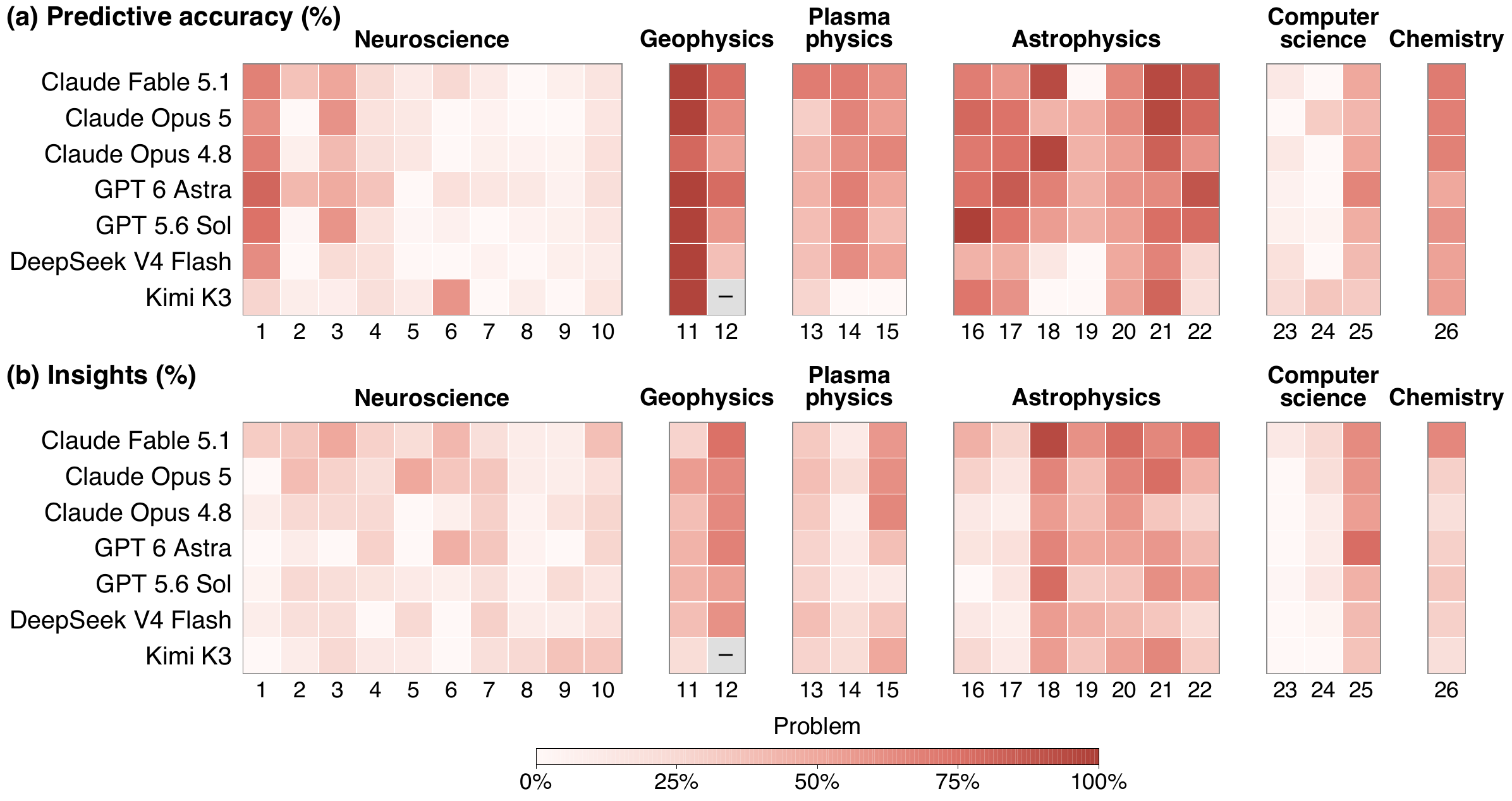}
    \caption{
       \textbf{PA and SI scores across domains.}
       Predictive accuracy (a) and SI scores (b) for seven AI agents on 26 problems across six domains. Each cell shows one agent's score on one problem, averaged over its instances using the same rules as Table~\ref{tab:main_results}. Both panels use the same color scale. Gray dashes mark the excluded Kimi K3 run on \textit{catchment properties}.
      %  \jy{will redo the plot, some scores are incorrects}
      % \zw{I don't understand these figures, make sure to provide analysis and informative captions.}
       }
    \label{fig:heatmap}
\end{figure}

Figure~\ref{fig:heatmap} compares PA and SI scores across the six domains.
On the problems in \ourbenchmark, both scores are relatively low in neuroscience and computer science.
Agents achieve higher PA in geophysics, plasma physics, and chemistry, but their SI scores often remain much lower.
Differences also emerge among agents with similar PA within a domain.
In astrophysics, the three Claude agents and two GPT agents achieve domain-average PA of 67.1\%--69.8\%, while their SI scores range from 33.6\% to 63.4\%.
These results show that agents with similar PA can still differ substantially in the scientific insights their mechanisms support.

\subsection{Does higher accuracy imply more insights?}
\label{app:pa_insights_stats}
We compare agents' PA and SI rankings within each problem using Kendall's $\tau_b$ \citep{kendall1945ties} (Table~\ref{tab:pa_insights_stats}).
We estimate 95\% confidence intervals from 20,000 bootstrap samples of whole problems \citep{efron1979bootstrap}.
The mean correlation is positive but limited ($\tau_b=0.20$, adjusted $p=0.008$).
After excluding pairs tied on either metric, the higher-PA agent scores lower on SI in 39.9\% of comparisons on average across problems.
The reversal rate remains similar when we use either judge's scores separately.
Predictive accuracy alone therefore does not reliably indicate how much scientific insight a mechanism supports.
% Generated by ahabench_plots/analyze_pa_insights.py.
% Bootstrap samples: 20000; blocked permutations: 100000.
% Reversal rates exclude pairs tied on either metric and give problems equal weight.
\begin{table}[htbp]
\centering
\begin{minipage}{0.75\textwidth}
\small
\setlength{\tabcolsep}{2pt}
\arrayrulecolor{black}
\begin{tabular*}{\linewidth}{@{\extracolsep{\fill}}lccc@{}}
\toprule
\textbf{Judge} & $\boldsymbol{\tau_b}$ & \textbf{Reversals (\%)} & $\boldsymbol{p}$ \\
\midrule
% \makecell[l]{Both judges\\(average)} & \makecell{0.20\\{\footnotesize[0.05, 0.34]}} & \makecell{39.6\\{\footnotesize[31.9, 47.4]}} & 0.014 \\
\makecell[l]{Both judges\\(average)} & \makecell{0.20\\{\footnotesize[0.07, 0.32]}} & \makecell{39.9\\{\footnotesize[33.2, 46.5]}} & 0.008 \\
\addlinespace[2pt]
% \makecell[l]{Claude Opus\\4.8} & \makecell{0.13\\{\footnotesize[0.00, 0.25]}} & \makecell{42.8\\{\footnotesize[35.8, 49.7]}} & 0.070 \\
\makecell[l]{Claude Opus\\4.8} & \makecell{0.15\\{\footnotesize[0.03, 0.26]}} & \makecell{41.8\\{\footnotesize[35.3, 48.4]}} & 0.054 \\
\addlinespace[2pt]
% GPT-5.6 Sol & \makecell{0.17\\{\footnotesize[0.03, 0.31]}} & \makecell{41.0\\{\footnotesize[33.4, 48.9]}} & 0.032 \\
GPT 5.6 Sol & \makecell{0.15\\{\footnotesize[0.01, 0.29]}} & \makecell{42.8\\{\footnotesize[34.3, 51.9]}} & 0.054 \\
\addlinespace[2pt]
\bottomrule
\end{tabular*}
\captionsetup{font=small,skip=4pt}
\caption{\textbf{PA and SI rankings.} Brackets show 95\% bootstrap confidence intervals; $p$-values are computed using 100,000 within-problem permutations under the null hypothesis of no association and adjusted across the three rows using Holm's method \citep{phipson2010permutation,holm1979simple}.}
\label{tab:pa_insights_stats}
\end{minipage}
\end{table}

\FloatBarrier

% D.3: Expert-supported failure categories, with original AI scientist agent evidence.
\clearpage
\subsection{Common failures of AI scientist agents}
\label{app:common_failures}
The ten categories below describe failures in the investigations and explanations of AI scientist agents.
Highlighted passages show the actions and claims relevant to each failure.

\paragraph{Missing tests of the target behavior.}
AI scientist agents investigate changes in internal activity without simulating or measuring the target behavior itself.
The missing test is whether the proposed mechanism actually reproduces the target behavior, including its timing and response to changing inputs.
\begin{figure}[H]
\centering
\includegraphics[page=1,width=\textwidth]{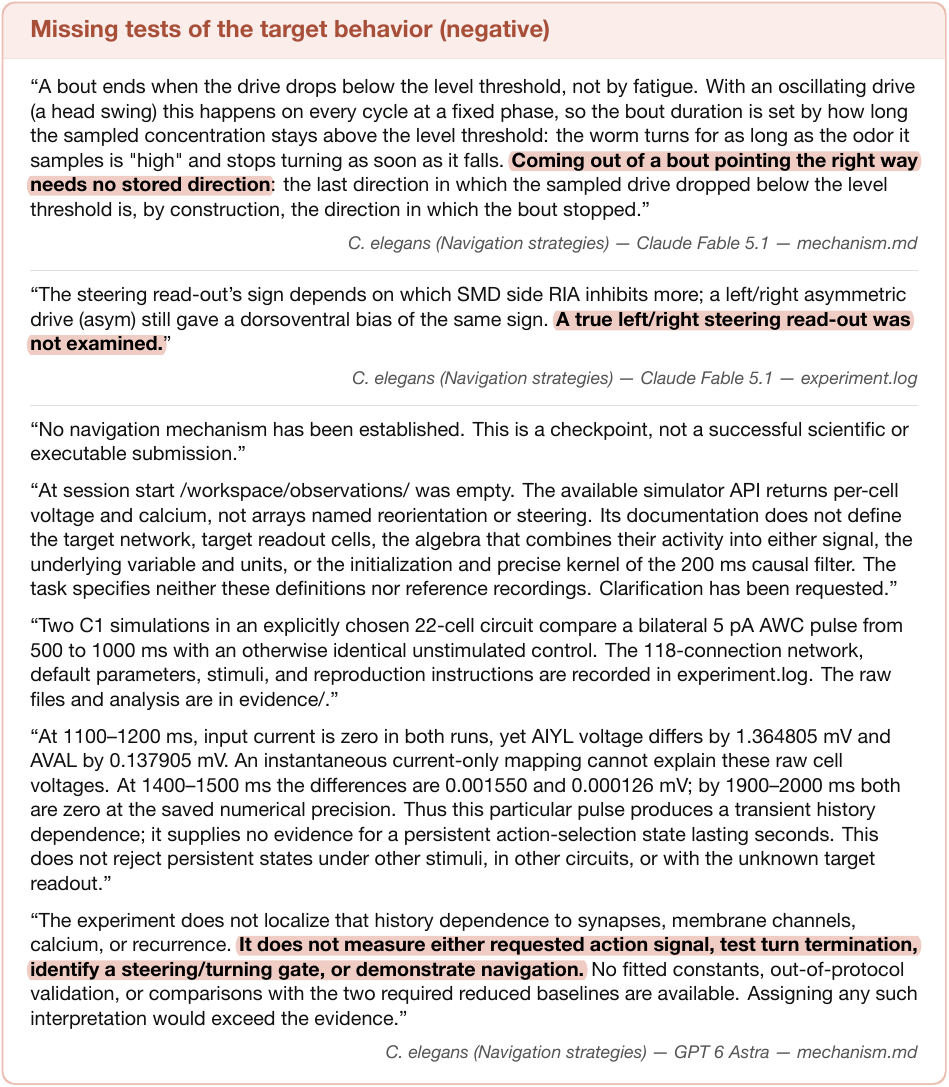}
\caption{\textbf{Missing tests of the target behavior.} Original outputs from AI scientist agents.}\label{fig:failure-missing-behavior-tests}
\end{figure}

\clearpage
\begin{figure}[H]
\centering
\ContinuedFloat
\includegraphics[page=2,width=\textwidth]{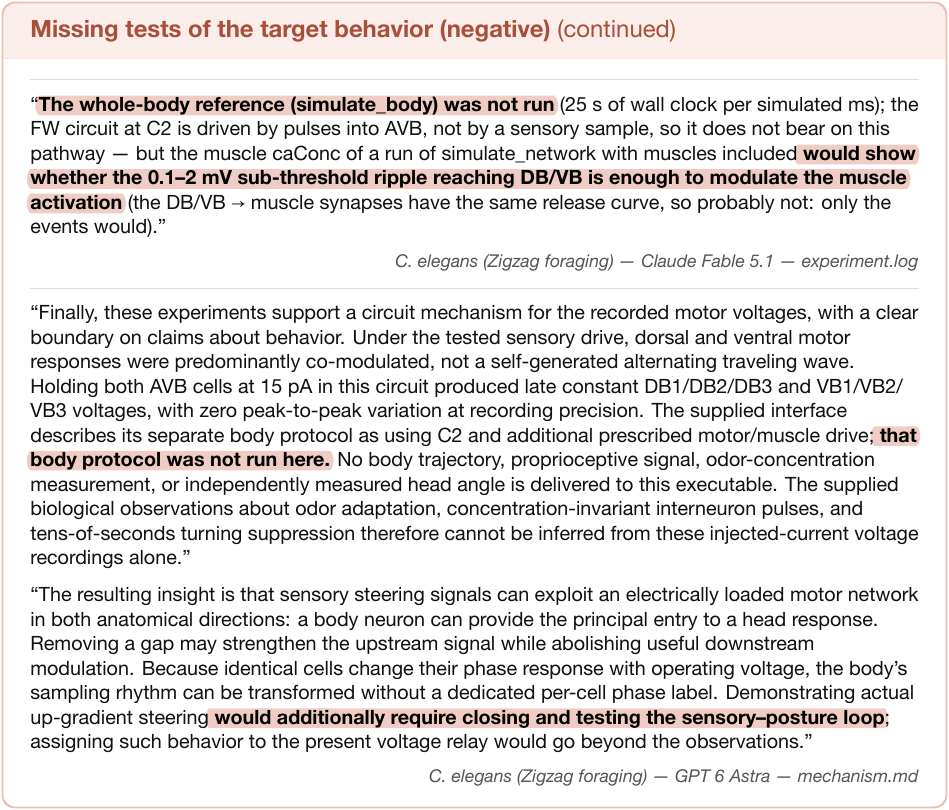}
\caption[]{\textbf{Missing tests of the target behavior} (continued).}
\end{figure}

\clearpage
\paragraph{Explaining only part of the target phenomenon.}
An explanation can account for how one component responds while leaving unexplained how components coordinate, how a state begins or ends, or how stored information changes an outcome.
AI scientist agents develop local descriptions but do not connect them into an account of the complete process they set out to explain.
\begin{figure}[H]
\centering
\includegraphics[page=1,width=\textwidth]{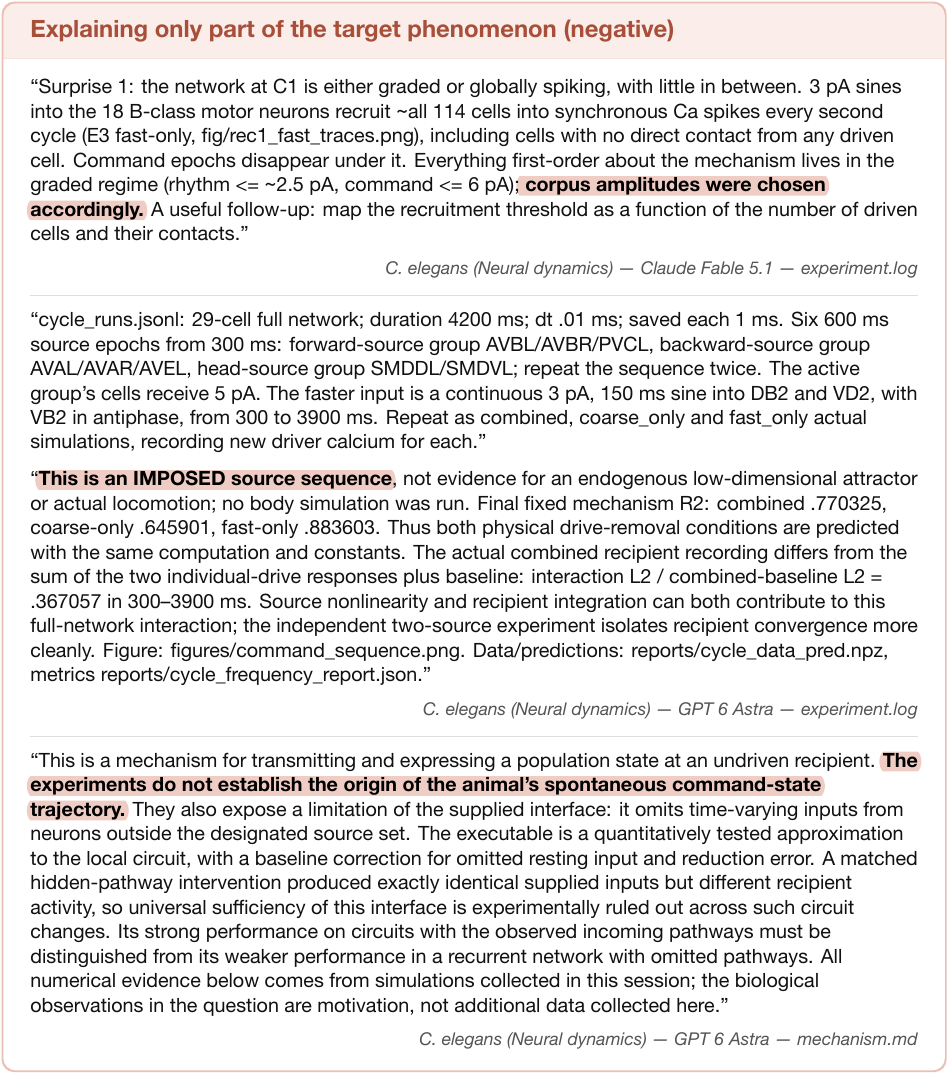}
\caption{\textbf{Explaining only part of the target phenomenon.} Original outputs from AI scientist agents.}\label{fig:failure-explaining-only-part-of-the-target-phenomenon}
\end{figure}

\clearpage
\begin{figure}[H]
\centering
\ContinuedFloat
\includegraphics[page=2,width=\textwidth]{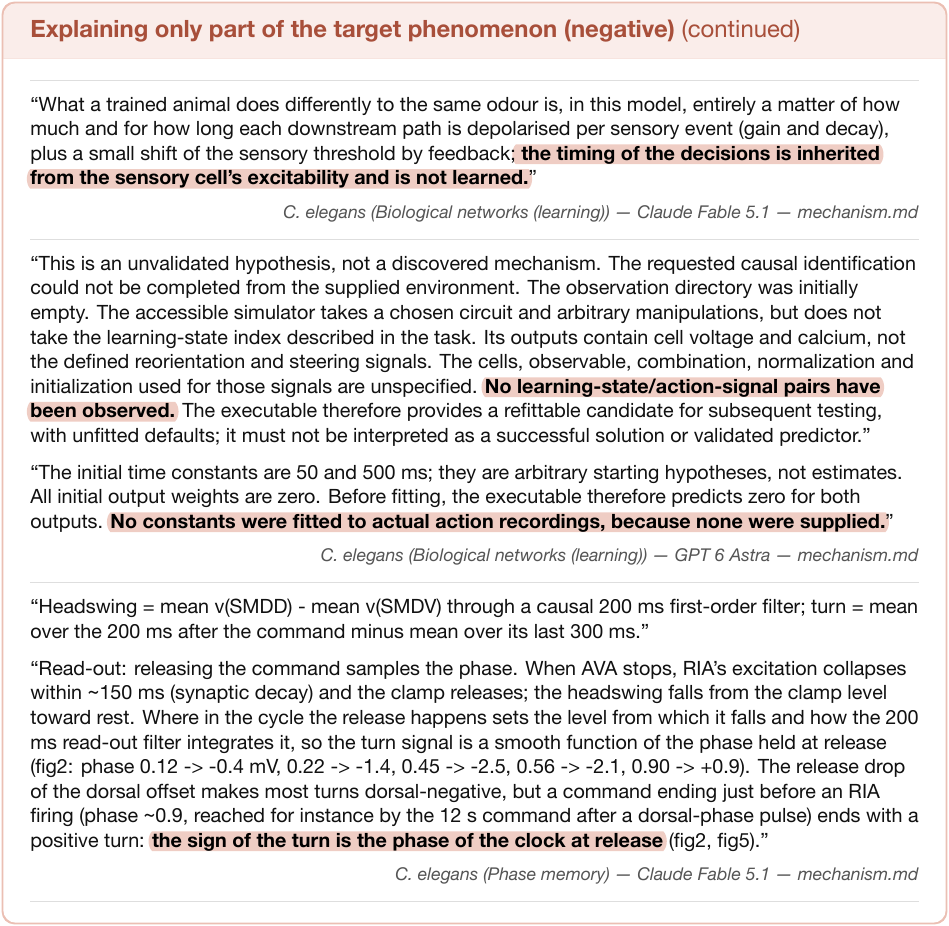}
\caption[]{\textbf{Explaining only part of the target phenomenon} (continued).}
\end{figure}

\clearpage
\begin{figure}[H]
\centering
\ContinuedFloat
\includegraphics[page=3,width=\textwidth]{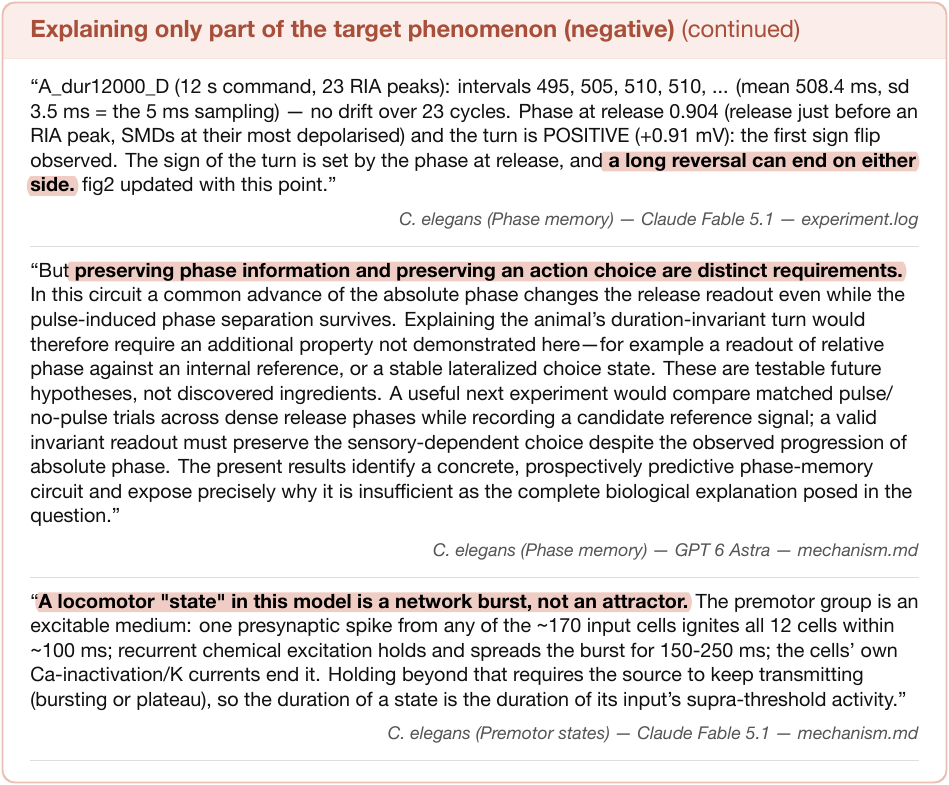}
\caption[]{\textbf{Explaining only part of the target phenomenon} (continued).}
\end{figure}

\clearpage
\begin{figure}[H]
\centering
\ContinuedFloat
\includegraphics[page=4,width=\textwidth]{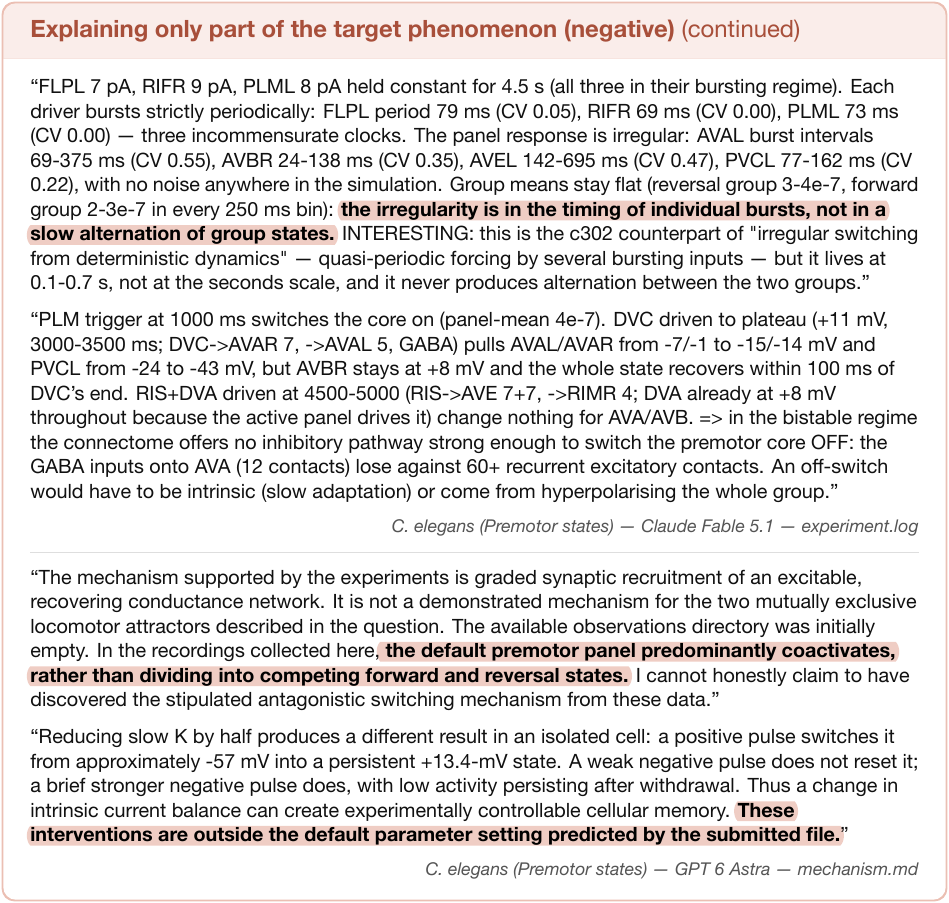}
\caption[]{\textbf{Explaining only part of the target phenomenon} (continued).}
\end{figure}

\clearpage
\paragraph{Prediction with limited explanation.}
An input–output rule can reproduce measured responses while leaving open how the system uses those responses or why its components are organized that way.
AI scientist agents give detailed accounts of how to compute the output, but provide limited explanations of the processes and functions that the predictions are meant to represent.
\begin{figure}[H]
\centering
\includegraphics[page=1,width=\textwidth]{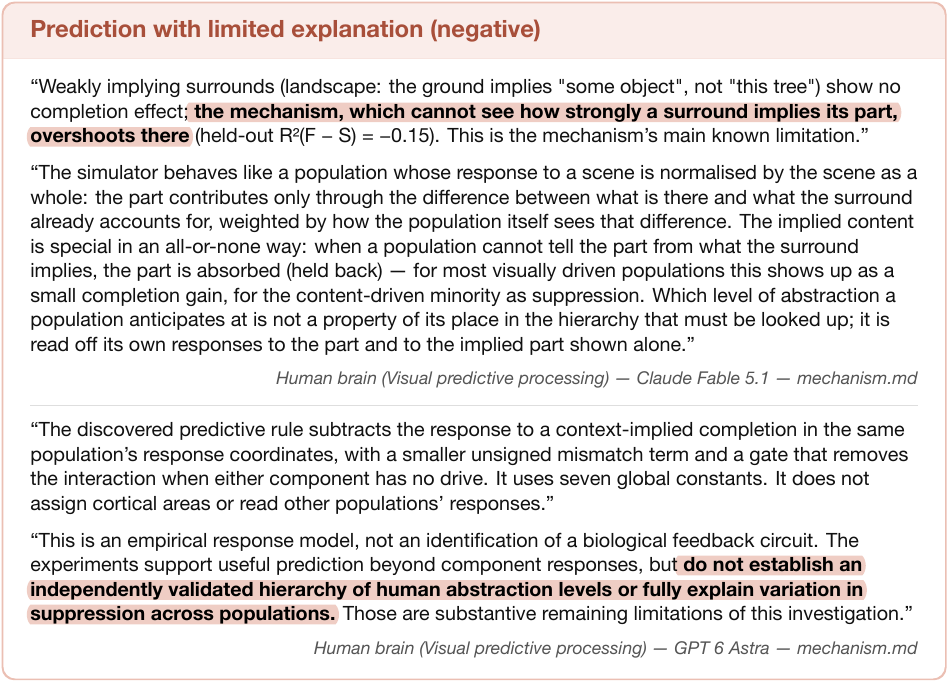}
\caption{\textbf{Prediction with limited explanation.} Original outputs from AI scientist agents.}\label{fig:failure-prediction-with-limited-explanation}
\end{figure}

\clearpage
\begin{figure}[H]
\centering
\ContinuedFloat
\includegraphics[page=2,width=\textwidth]{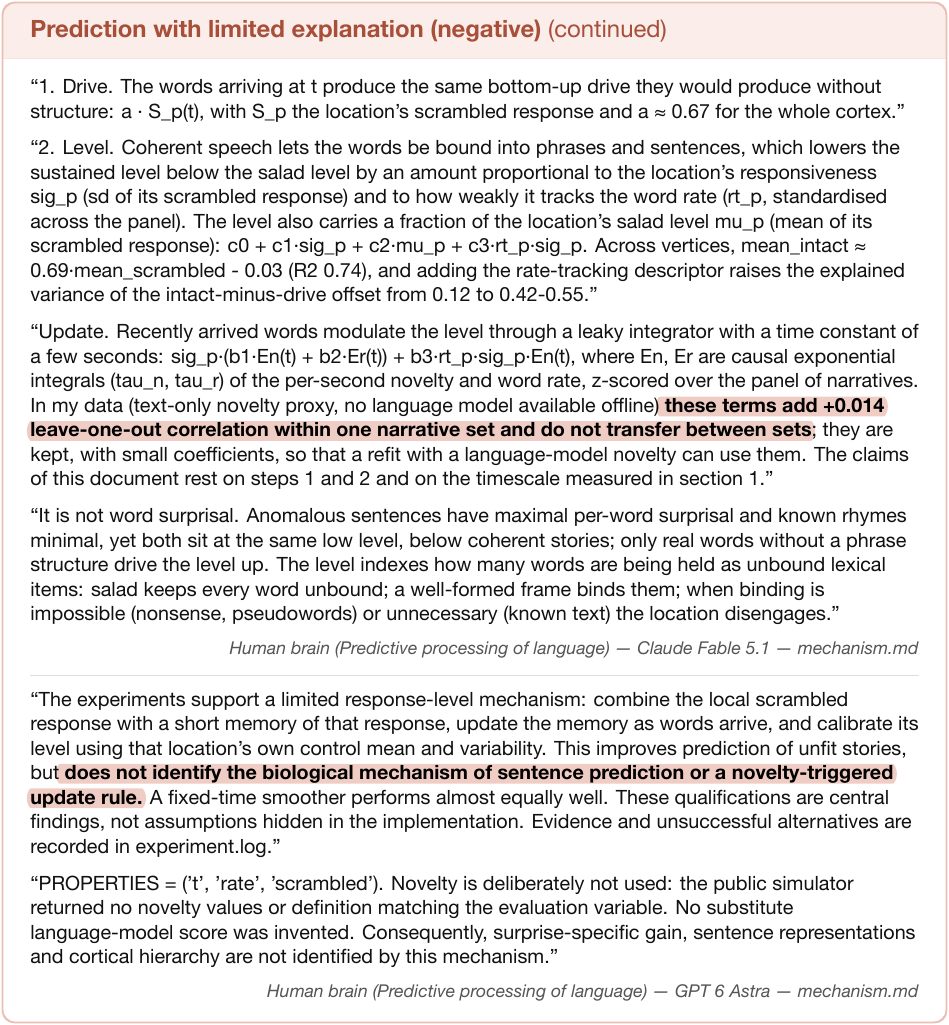}
\caption[]{\textbf{Prediction with limited explanation} (continued).}
\end{figure}

\clearpage
\begin{figure}[H]
\centering
\ContinuedFloat
\includegraphics[page=3,width=\textwidth]{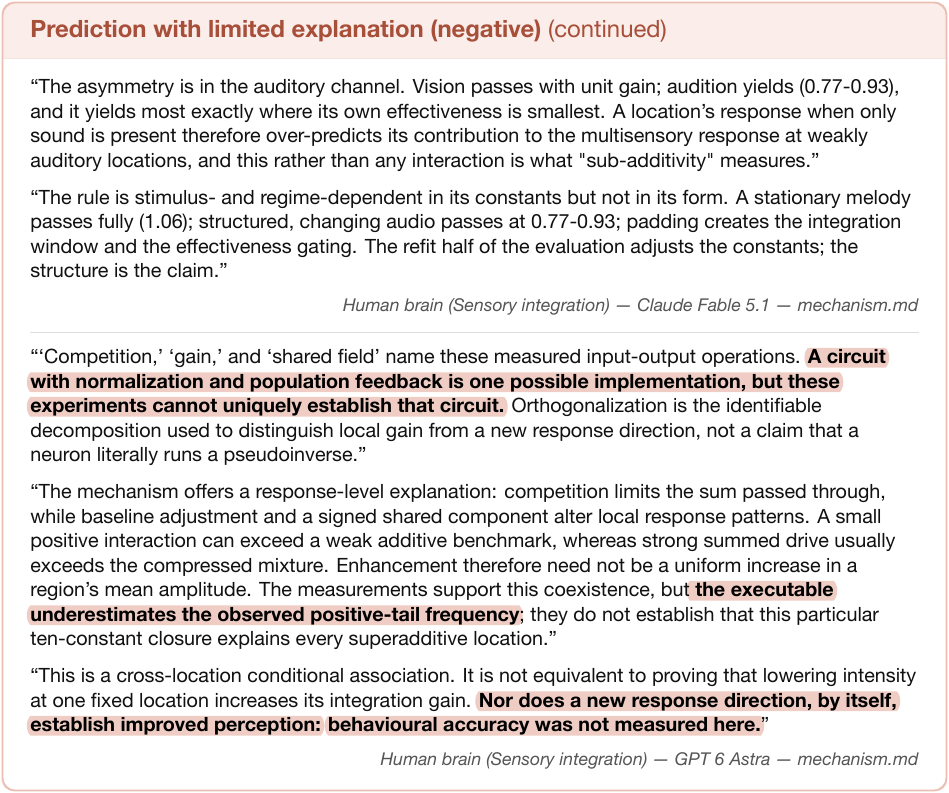}
\caption[]{\textbf{Prediction with limited explanation} (continued).}
\end{figure}

\clearpage
\begin{figure}[H]
\centering
\ContinuedFloat
\includegraphics[page=4,width=\textwidth]{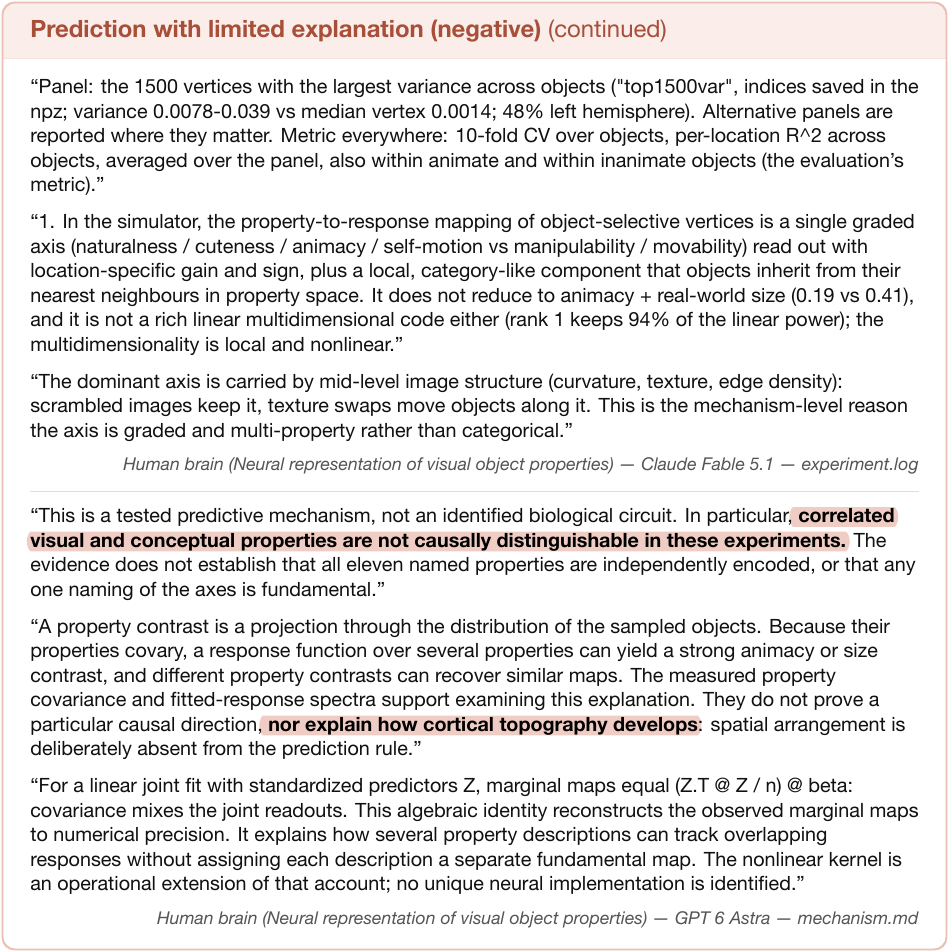}
\caption[]{\textbf{Prediction with limited explanation} (continued).}
\end{figure}

\clearpage
\paragraph{Unclear explanations.}
Human scientists need to know what was changed in an experiment, what was observed, and how the observation supports the claim.
AI scientist agents omit parts of this chain, use unexplained shorthand, or bury the main findings among implementation details.
Both the course of the investigation and the proposed mechanism then become difficult to reconstruct from the report.
\begin{figure}[H]
\centering
\includegraphics[page=1,width=\textwidth]{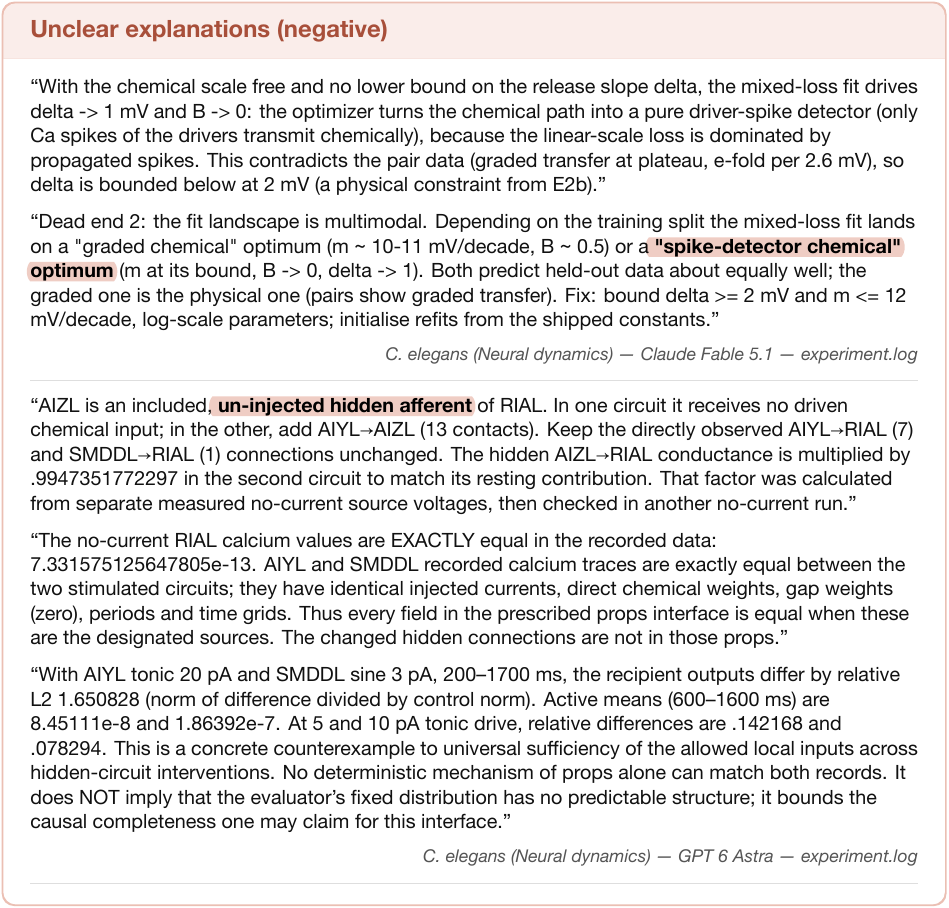}
\caption{\textbf{Unclear explanations.} Original outputs from AI scientist agents.}\label{fig:failure-unclear-explanations}
\end{figure}

\clearpage
\begin{figure}[H]
\centering
\ContinuedFloat
\includegraphics[page=2,width=\textwidth]{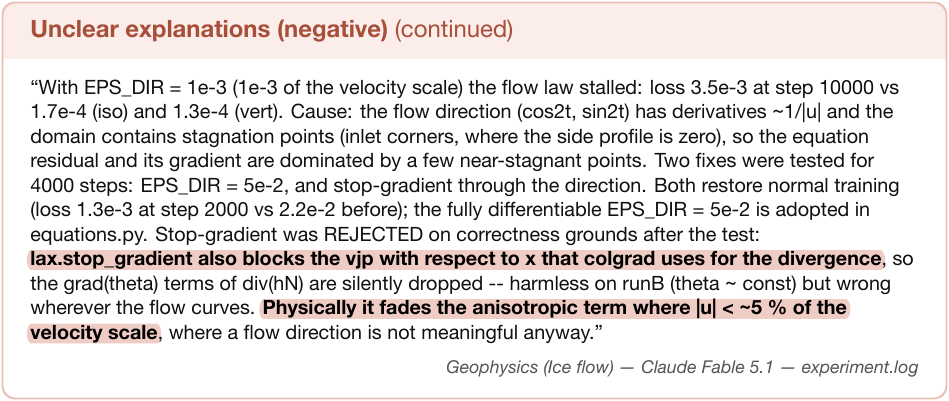}
\caption[]{\textbf{Unclear explanations} (continued).}
\end{figure}

\clearpage
\paragraph{Overgeneralization.}
AI scientist agents turn an observation made under limited conditions or within a limited time window into a claim about what the system always or never does.
A response that is absent in the tested window is treated as absent altogether, even though the experiments have not excluded it on other timescales or under other conditions.
\begin{figure}[H]
\centering
\includegraphics[page=1,width=\textwidth]{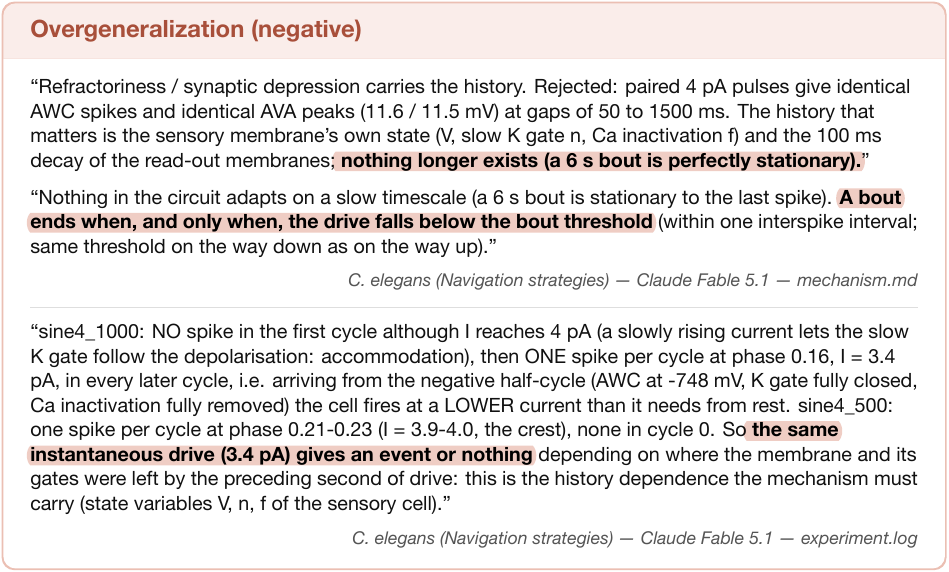}
\caption{\textbf{Overgeneralization.} Original outputs from AI scientist agents.}\label{fig:failure-overgeneralization}
\end{figure}

\clearpage
\paragraph{Failing to identify and reduce uncertainty.}
AI scientist agents do not reliably identify which parts of an explanation are established by evidence and which still depend on their own choices.
They can assume that a chosen measurement captures the target phenomenon or that a restricted set of hypotheses covers the relevant possibilities, without designing experiments to test those assumptions.
Even when an unresolved question appears in the report, it may remain a caveat rather than guide the next experiment.
\begin{figure}[H]
\centering
\includegraphics[page=1,width=\textwidth]{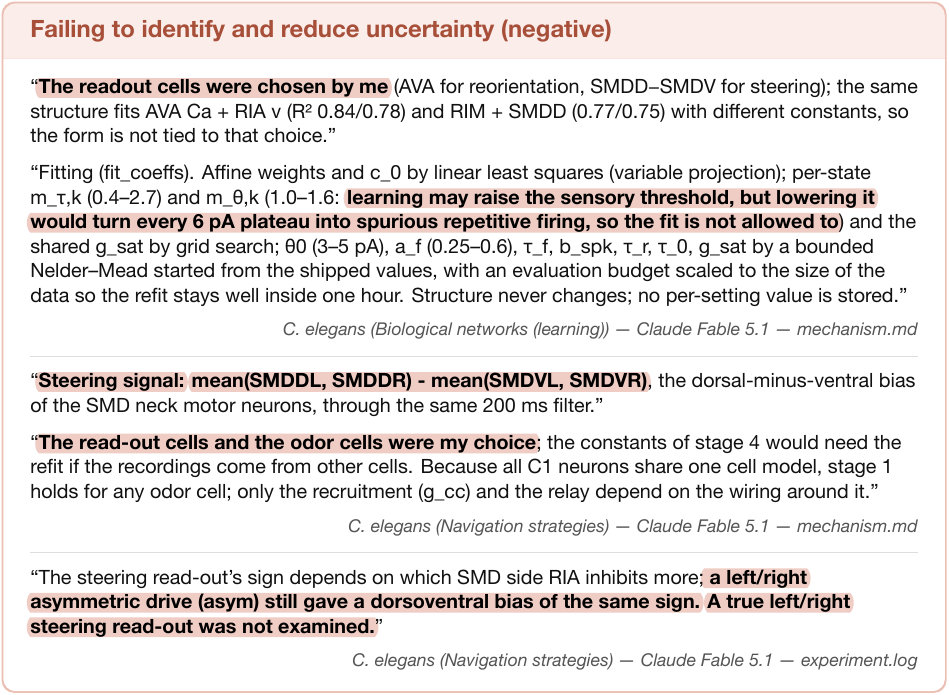}
\caption{\textbf{Failing to identify and reduce uncertainty.} Original outputs from AI scientist agents.}\label{fig:failure-conclusions-limited-by-chosen-assumptions}
\end{figure}

\clearpage
\paragraph{Treating task optimization as mechanism discovery.}
AI scientist agents turn the investigation into a search for better-fitting parameters or a more accurate recovery procedure.
These improvements show how to solve the task more effectively, but leave open why the system shows the effect that motivated the task (for example, why fitting systematically favors particular parameter values).
The failure is treating the improved solution as the main discovery without investigating that remaining scientific question.
\begin{figure}[H]
\centering
\includegraphics[page=1,width=\textwidth]{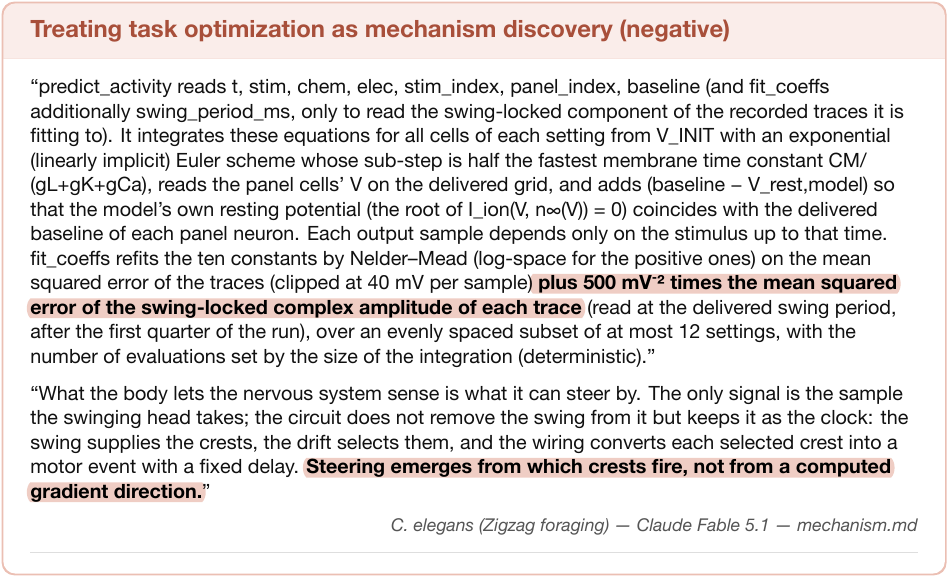}
\caption{\textbf{Treating task optimization as mechanism discovery.} Original outputs from AI scientist agents.}\label{fig:failure-treating-good-fit-as-mechanism}
\end{figure}

\clearpage
\begin{figure}[H]
\centering
\ContinuedFloat
\includegraphics[page=2,width=\textwidth]{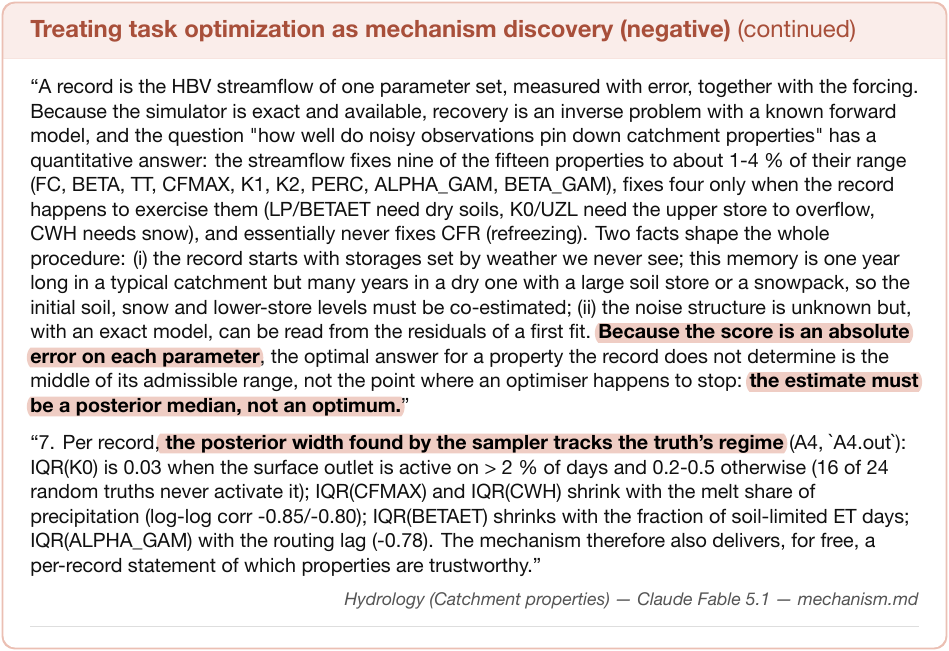}
\caption[]{\textbf{Treating task optimization as mechanism discovery} (continued).}
\end{figure}

\clearpage
\begin{figure}[H]
\centering
\ContinuedFloat
\includegraphics[page=3,width=\textwidth]{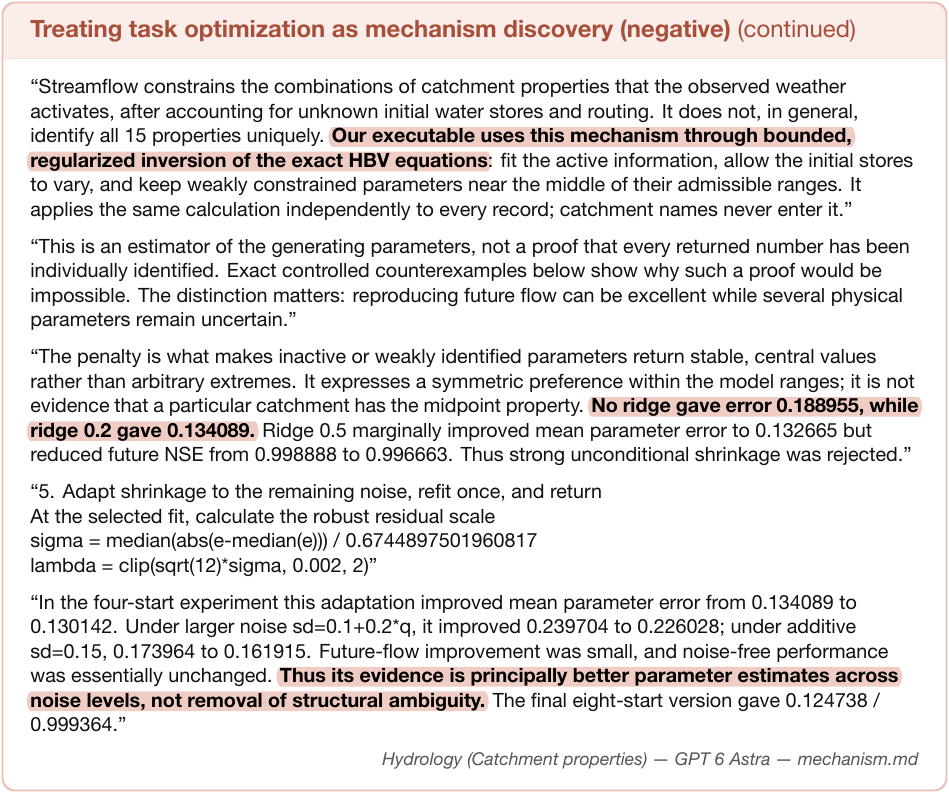}
\caption[]{\textbf{Treating task optimization as mechanism discovery} (continued).}
\end{figure}

\clearpage
\paragraph{Interesting observations without further exploration.}
A useful observation can suggest a new experiment: change the condition that appears to control an effect, or test a different regime.
AI scientist agents record the observation but do not run this experiment.
Agents thus miss the opportunity to pursue the research direction opened by the finding, even when they have run many other experiments.
\begin{figure}[H]
\centering
\includegraphics[page=1,width=\textwidth]{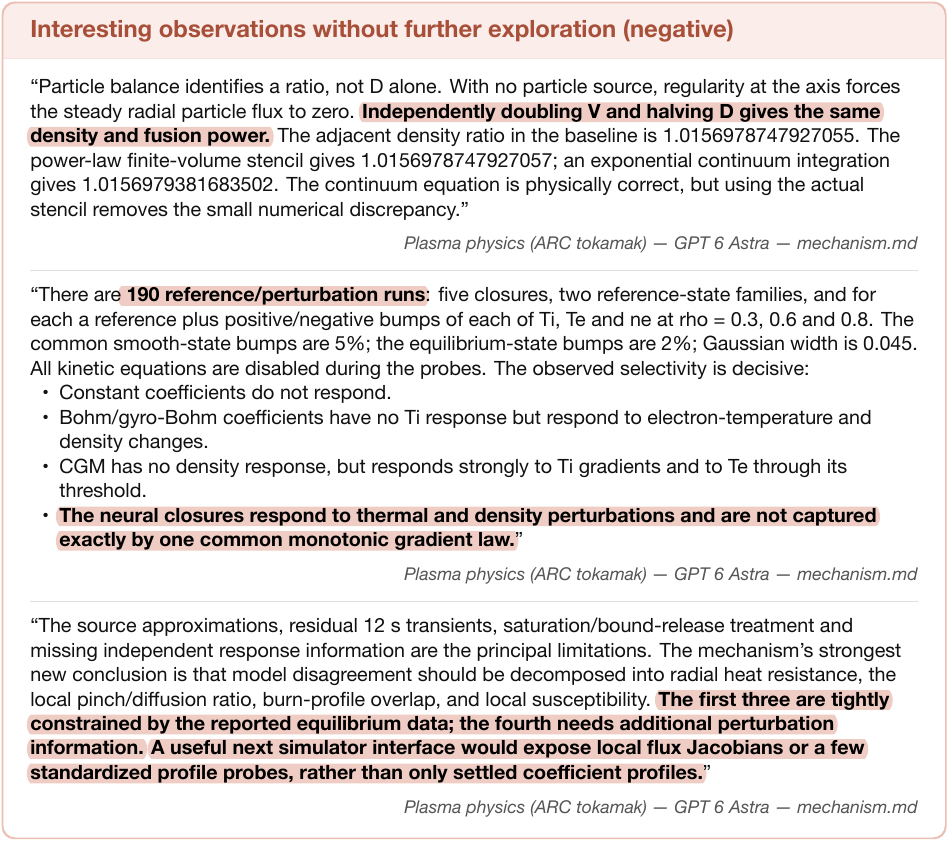}
\caption{\textbf{Interesting observations without further exploration.} Original outputs from AI scientist agents.}\label{fig:failure-interesting-observations-without-further-exploration}
\end{figure}

\clearpage
\paragraph{Conflicting results and conclusions.}
An experiment log records an effect under a particular condition, while the mechanism summary says that the effect was not found or reports it under a different condition.
AI scientist agents leave these statements inconsistent, without correcting the summary or explaining the difference.
The report therefore gives incompatible accounts of what the experiment established.
\begin{figure}[H]
\centering
\includegraphics[page=1,width=\textwidth]{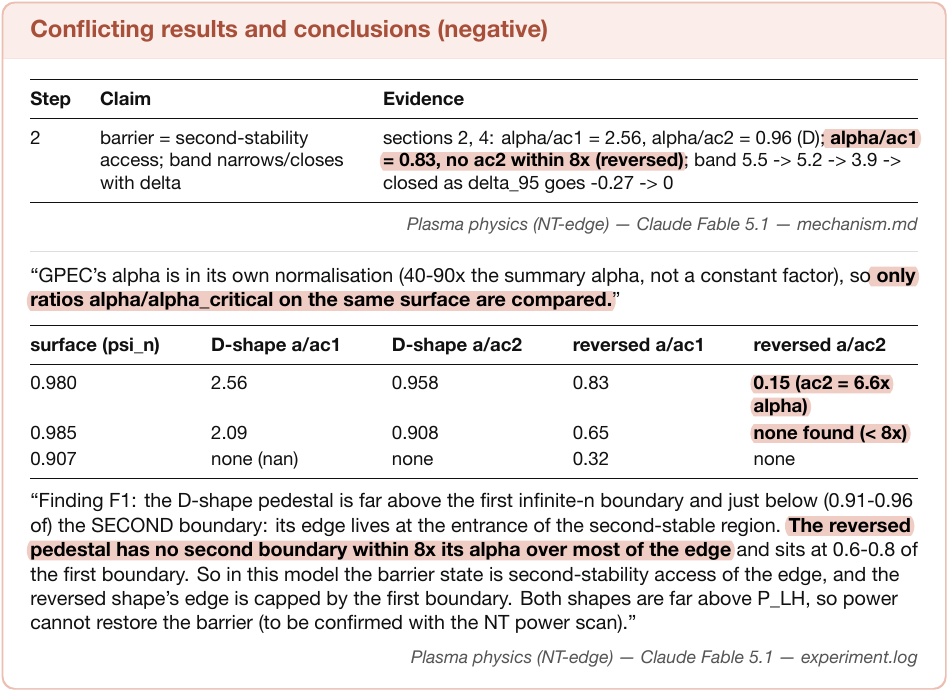}
\caption{\textbf{Conflicting results and conclusions.} Original outputs from AI scientist agents.}\label{fig:failure-conflicting-results-and-conclusions}
\end{figure}

\clearpage
\paragraph{Risk of circular validation.}
AI scientist agents can create test materials, assign descriptive properties to them, and then fit models and evaluate their predictions based on those same choices.
If the materials and their assigned properties were designed to vary together, successful prediction may recover a relationship introduced during construction.
Such agreement needs an independent check before it can establish that the relationship explains the system being studied.
\begin{figure}[H]
\centering
\includegraphics[page=1,width=\textwidth]{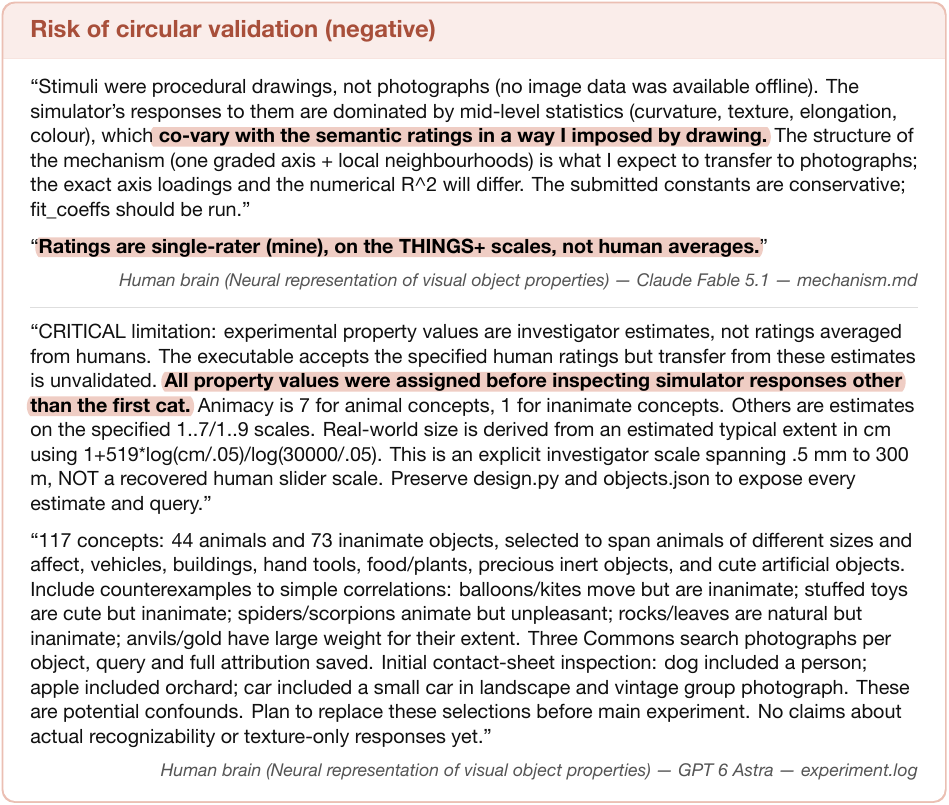}
\caption{\textbf{Risk of circular validation.} Original outputs from AI scientist agents.}\label{fig:failure-risk-of-circular-validation}
\end{figure}

\FloatBarrier\clearpage

% D.4: Expert-supported positive behaviors, with representative original evidence.
\subsection{Common positive behaviors of AI scientist agents}
\label{app:common_positives}
The four categories below describe useful experimental practices used by AI scientist agents and informative observations they make.
Highlighted passages show how these behaviors appear in the original outputs.

\paragraph{Using controls to rule out alternative explanations.}
When an intervention changes several quantities at once, the resulting response may have more than one explanation.
AI scientist agents distinguish between these explanations by holding a competing factor fixed or comparing inputs designed to isolate its contribution.
A difference that survives the control can rule out that factor as the sole explanation of the effect.
\begin{figure}[H]
\centering
\includegraphics[page=1,width=\textwidth]{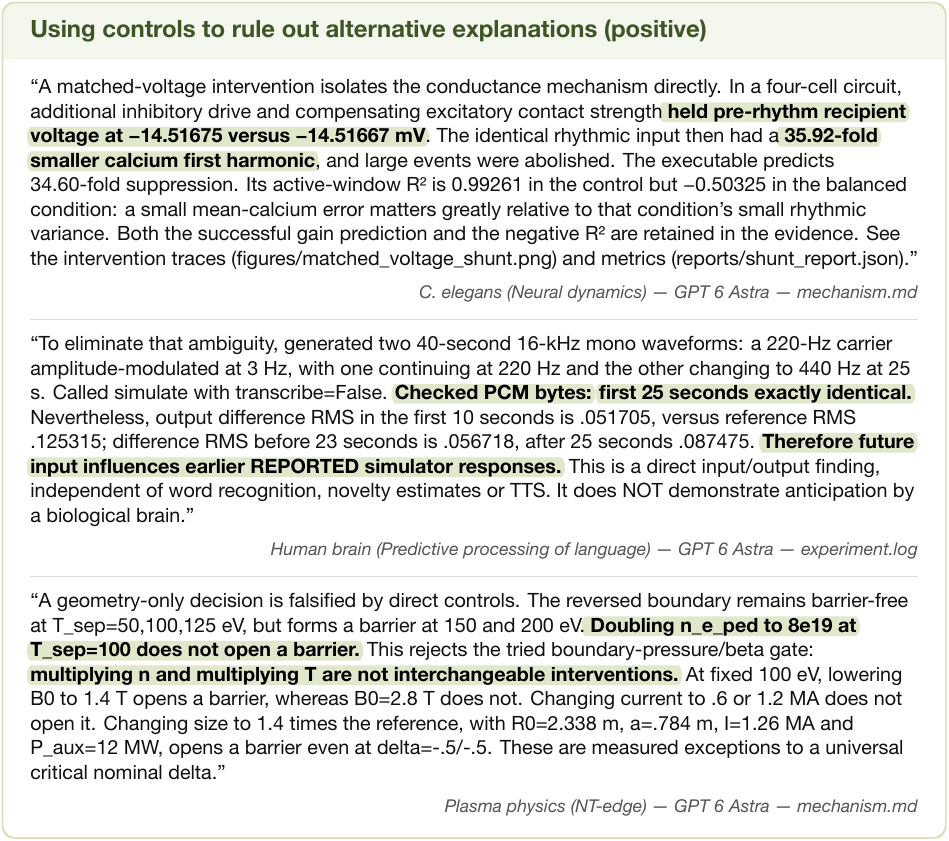}
\caption{\textbf{Using controls to rule out alternative explanations.} Original outputs from AI scientist agents.}\label{fig:positive-control-experiments}
\end{figure}

\clearpage
\paragraph{Testing which model better predicts the observations.}
Comparing models with and without a proposed component tests whether that component adds predictive value.
AI scientist agents compare the models on new observations and examine uncertainty in the performance difference.
The result can justify removing a term whose apparent benefit disappears, as well as retaining one supported by the comparison.
\begin{figure}[H]
\centering
\includegraphics[page=1,width=\textwidth]{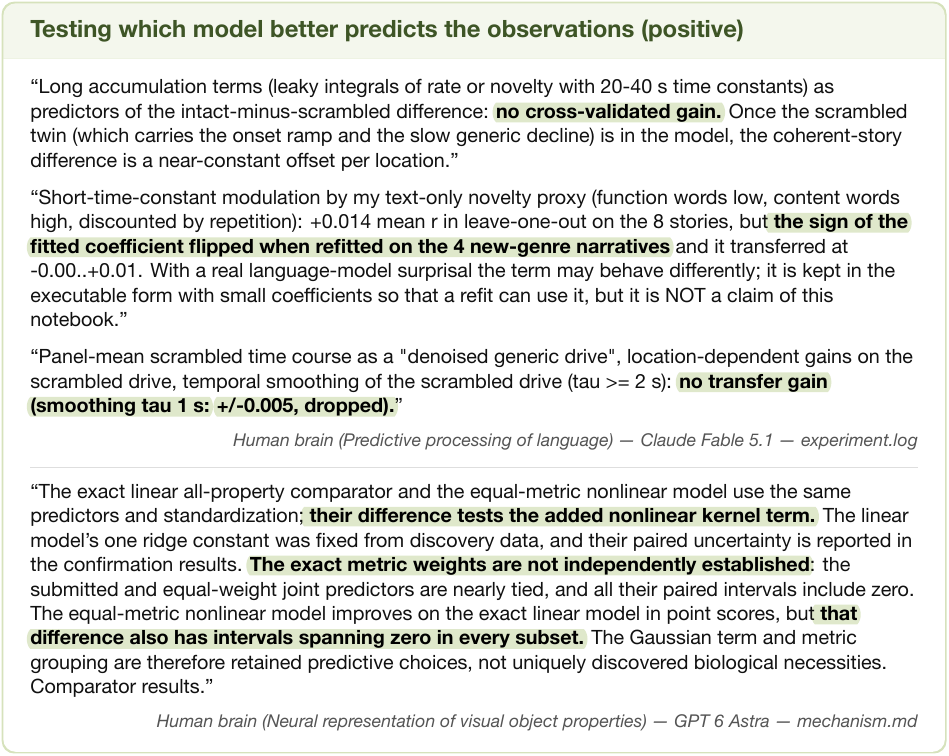}
\caption{\textbf{Testing which model better predicts the observations.} Original outputs from AI scientist agents.}\label{fig:positive-model-comparison}
\end{figure}

\clearpage
\paragraph{Revising models after experimental feedback.}
A failed prediction can expose an incorrect interpretation of an effect or a weakness in the fitting procedure.
AI scientist agents respond by changing the working explanation or model and setting up a new comparison to test the revision.
\begin{figure}[H]
\centering
\includegraphics[page=1,width=\textwidth]{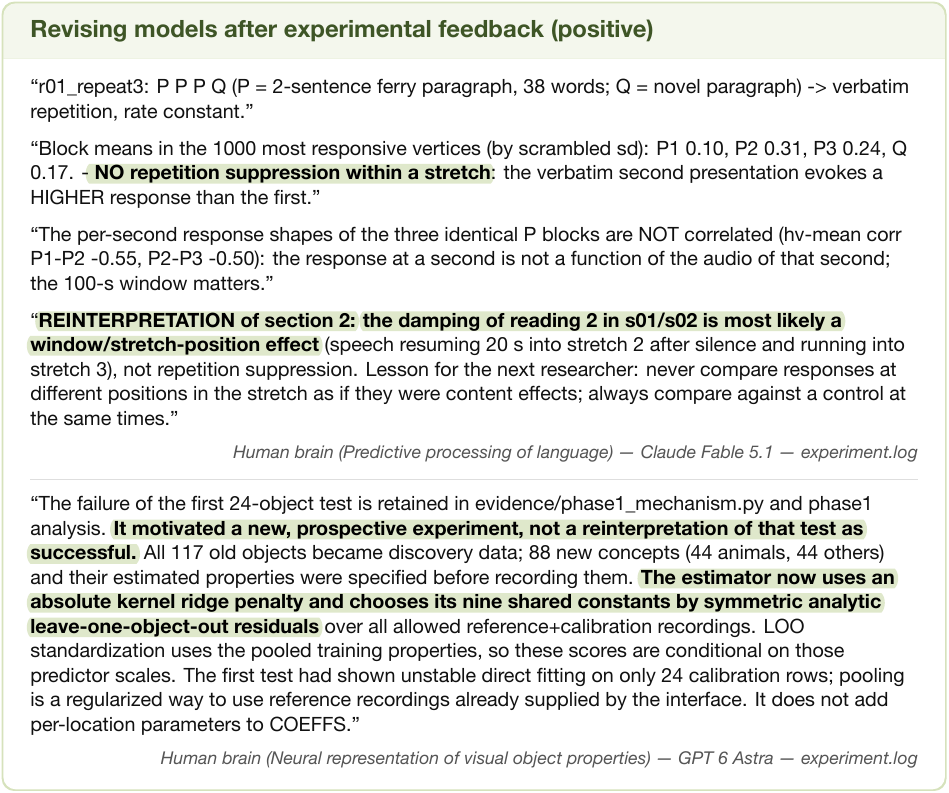}
\caption{\textbf{Revising models after experimental feedback.} Original outputs from AI scientist agents.}\label{fig:positive-revising-models-after-experimental-feedback}
\end{figure}

\clearpage
\paragraph{Informative observations.}
Some observations reveal distinctions that an overall fit score hides: two measurements need not change together, an intervention can shift the timing of a response without changing its shape, and some parameters remain poorly constrained when their associated process is inactive.
AI scientist agents recognize these patterns and use them to describe what the measurements reveal about the system and under which conditions these findings hold.
\begin{figure}[H]
\centering
\includegraphics[page=1,width=\textwidth]{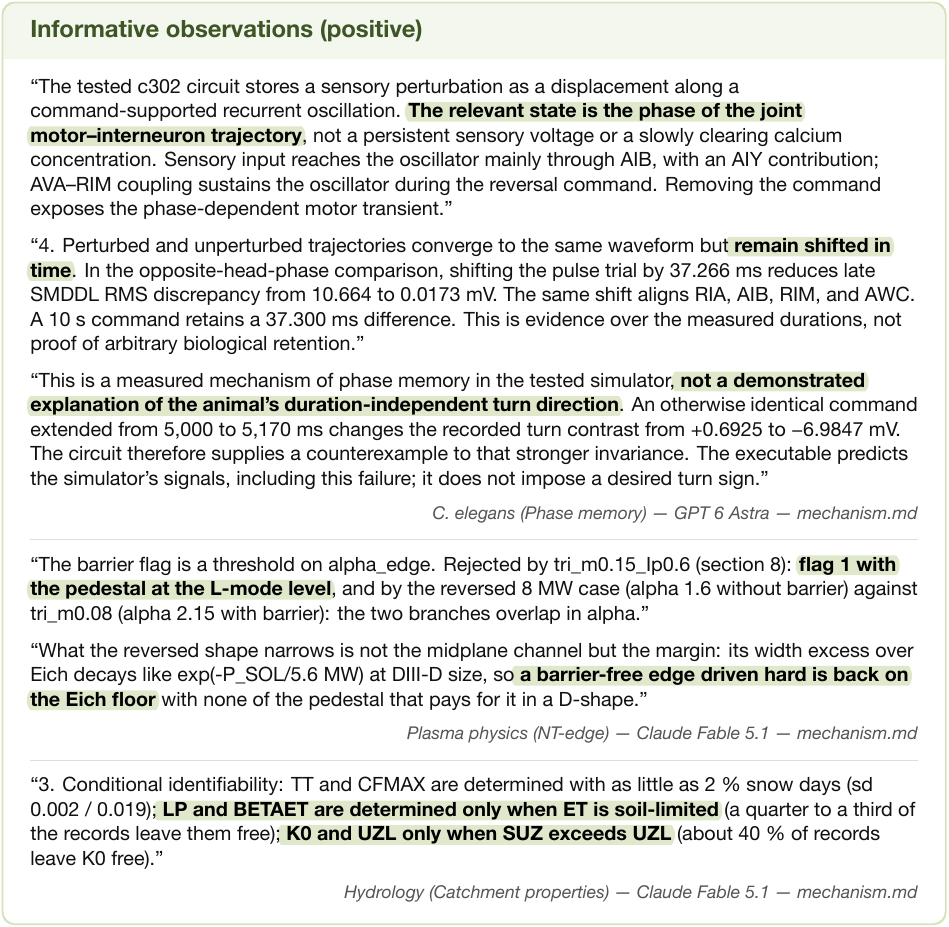}
\caption{\textbf{Informative observations.} Original outputs from AI scientist agents.}\label{fig:positive-informative-observations}
\end{figure}

\FloatBarrier\clearpage

% Shared appendix for domain environment interfaces; original E--J sections are preserved below.
\section{Environment interfaces}
\label{app:environment_interfaces}

\subsection{Neuroscience}
\label{app:neuroscience_environments}
\paragraph{c302.}
c302~\citep{gleeson2018c302} builds NeuroML models of \textit{C. elegans} neurons and muscles from measured connectivity, with different levels of biophysical detail.
The six \textit{C. elegans} problems share a CLI through which the scientist agent can inspect the connectome, choose the neurons and muscles to include, and simulate their activity.
In experiments, the scientist agent can inject current into selected neurons, remove connections, or change synaptic weights and cellular parameters; the resulting activity traces record the effects of these interventions on the circuit.

\paragraph{TRIBE v2.}
For the four human brain problems, TRIBE v2~\citep{d2026foundation} predicts functional magnetic resonance imaging (fMRI) responses to stimuli constructed by the scientist agent.
The agent submits video, audio, or text files through a shared CLI, with text converted to speech before processing.
The model returns predicted activity at 20,484 locations across the cortical surface over the duration of the stimulus, and the interface also provides the cortical geometry for locating these responses.

\subsection{Geophysics}
\label{app:geophysics_environments}
\paragraph{FEniCSx.}
FEniCSx is an open-source computing framework for solving partial differential equations with the finite element method~\citep{baratta2023dolfinx}.
In the ice shelf flow laws problem, the scientist agent writes and runs its own solver using our FEniCSx installation and instructions for coupling the two-dimensional shallow-shelf approximation (SSA) momentum equations with mass conservation.
The equations, viscosity, boundary conditions, and shelf geometry can be changed to generate velocity and thickness fields under different physical assumptions.

\paragraph{HBV.}
We construct the ground truth for catchment properties with HBV~\citep{bergstrom1992hbv}, a conceptual model of water storage and movement through snow, soil, and groundwater.
We choose known values for its 15 parameters and drive the model with daily precipitation, air temperature, and potential evapotranspiration from CAMELS-US~\citep{addor2017camels}.
To represent measurement uncertainty in the streamflow observations, we add noise to the simulated streamflow while retaining the generating parameters and noise-free streamflow as ground truth.
The scientist agent accesses the same model through a CLI, where it can select catchments, parameter values, and initial storage states and inspect the resulting streamflow, internal storages, and fluxes.

\subsection{Plasma Physics}
\label{app:plasma_environments}
\paragraph{TORAX.}
The ARC tokamak and negative triangularity core-only problems share an interface to TORAX~\citep{citrin2024torax}, a differentiable simulator of tokamak plasma transport.
TORAX solves coupled one-dimensional equations for ion and electron heat transport, particle transport, and current diffusion.
Through the CLI, the scientist agent chooses the magnetic geometry, transport model, heating sources, and boundary conditions and records the resulting evolution of plasma profiles and global quantities.
For the core-only problem, we also supply a library of matched magnetic equilibria with different triangularities so that plasma shape can be varied across runs.

\paragraph{TORAX, TokaMaker, and GPEC.}
The negative triangularity edge environment combines TORAX with TokaMaker~\citep{hansen2024tokamaker} and the Generalized Perturbed Equilibrium Code (GPEC)~\citep{glasser2016direct}.
TokaMaker solves the free-boundary Grad--Shafranov equation for the magnetic equilibrium, and GPEC calculates its ideal ballooning stability boundaries.
Our coupled model uses these stability calculations to determine the edge conditions for TORAX transport simulations.
The scientist agent can vary the plasma shape and operating conditions through the CLI, run the equilibrium, stability, and transport calculations together or individually, and inspect the resulting equilibria and plasma profiles.

\subsection{Astrophysics}
\label{app:astrophysics_environments}
\paragraph{CAMELS.}
CAMELS~\citep{villaescusa2021camels} contains precomputed galaxy-formation simulations with different cosmological parameters and models of stellar and black hole feedback.
We draw on suites that vary parameters individually or jointly, as well as runs with matched initial conditions across galaxy-formation models.
For the single galaxy, galaxy environment, and gravitational wave problems, the scientist agent selects and compares simulations through a common CLI that retrieves their parameters, halo and galaxy catalogs, particle snapshots, and available matter power spectra.

\paragraph{The DREAMS Project.}
The DREAMS Project~\citep{rose2025introducing} uses cosmological simulations and high-resolution simulations of Milky Way-mass halos to study the effects of dark matter physics and astrophysical feedback on galaxy formation.
For the Milky Way and baryonic feedback problems, we use cold dark matter simulations with hydrodynamic and gravity-only runs sharing the same initial conditions.
The WDM merge tree and WDM population problems use warm dark matter suites that vary the particle mass and feedback parameters.
A shared CLI retrieves simulation parameters and halo catalogs, with particle snapshots or merger trees available in the relevant suites.

\subsection{Computer Science}
\label{app:computer_science_environments}
\paragraph{Open-weight language models.}
Stated belief shift and subliminal learning share an interface for LoRA~\citep{hu2022lora} fine-tuning of open-weight language models on subsets of the supplied training data.
The scientist agent can remove selected examples and repeat training to test whether they are responsible for the observed change in beliefs or preferences.
Circuit discovery uses TransformerLens~\citep{nanda2022transformerlens} to test candidate circuits by replacing the activations of selected model components and measuring the change in the model's output.

\subsection{Chemistry}
\label{app:chemistry_environments}
\paragraph{AIMNet2-rxn.}
AIMNet2-rxn~\citep{anstine2025aimnet2} is a machine-learned potential for organic reactions, trained on density functional theory (DFT) calculations at the $\omega$B97M/def2-TZVPP level for neutral, closed-shell structures containing H, C, N, and O.
It returns potential energies and atomic forces, which our CLI uses to optimize geometries and search for transition states along candidate reaction paths.
Vibrational analysis checks whether a candidate is a saddle point, and intrinsic reaction coordinate calculations check whether it connects the intended reactants and products.
The scientist agent uses the resulting paths and energy differences to identify elementary steps and calculate activation barriers and reaction energies.

\paragraph{Sanity check.}
To check that the catalyst ground truth computed with AIMNet2-rxn is physically reasonable, we compare AIMNet2-rxn with published DFT calculations for three organocatalytic reactions: the proline-catalyzed Hajos--Parrish--Eder--Sauer--Wiechert cyclization~\citep{bahmanyar2001origin}, the proline-catalyzed aldol reaction of cyclohexanone and benzaldehyde~\citep{armstrong2014houk}, and the TBD-catalyzed ring-opening of lactide by methanol~\citep{chuma2008reaction}.
Starting from the published geometries, we check relative energies, optimized transition-state structures, and the ordering of competing reaction paths.
Of 30 reference transition states, 29 remain first-order saddle points after optimization, with heavy-atom root-mean-square deviations of 0.07--0.48~\AA{} from the DFT structures.
For the 16 aldol transition states, relative energies differ from the DFT values by 1.85~kcal/mol on average and 3.51~kcal/mol at most. The two methods' rankings have a Spearman correlation of 0.79, and both methods identify the same lowest-energy transition state.
For lactide, AIMNet2-rxn also favors the hydrogen-bonded route over the covalent route by 13.9~kcal/mol, compared with 11.7~kcal/mol in the reference.
However, barriers measured from reactant minima are 1.9--8.6~kcal/mol higher than the reference values, while the energy separation between two Hajos--Parrish transition states narrows from 0.18 to 0.01~kcal/mol.
Because these comparisons use different DFT approximations, discrepancies cannot be attributed entirely to AIMNet2-rxn.

\FloatBarrier\clearpage
\section{Problem overview}
\label{app:problem_overview}
\begin{table}[H]
\centering
\small
\setlength{\tabcolsep}{6pt}
\renewcommand{\arraystretch}{1.18}
\definecolor{problemneuro}{HTML}{E8856F}
\definecolor{problemgeo}{HTML}{7B9CB1}
\definecolor{problemplasma}{HTML}{F2AE75}
\definecolor{problemastro}{HTML}{B7587D}
\definecolor{problemcs}{HTML}{B0CBB1}
\definecolor{problemchem}{HTML}{CBC7B9}
\newcommand{\problemdomainbadge}[2]{%
  \begingroup
  \setlength{\fboxsep}{2pt}%
  \colorbox{#1!25}{\strut\textbf{#2}}%
  \endgroup
}
\begin{tabular}{@{}ll>{\itshape}l@{}}
\toprule
\textbf{Scientific domains} & \textbf{Problem names} & {\upshape\textbf{Sources}} \\
\midrule
\multirow{10}{*}{\problemdomainbadge{problemneuro}{Neuroscience}} & Premotor states & \citealp{morrison2025premotor} \\
& Zigzag foraging & \citealp{zhao2024integrative} \\
& Biological networks & \citealp{chen2025navigation} \\
& Neural dynamics & \citealp{fieseler2025intrinsic} \\
& Navigation strategies & \citealp{chen2026stateswitching} \\
& Phase memory & \citealp{dunn2025memory} \\
& Sensory integration & \citealp{buhmann2026inverted} \\
& Visual predictive processing & \citealp{scheurer2026spatial} \\
& Predictive processing of language & \citealp{zhou2026hierarchical} \\
& Neural representation & \citealp{stoinski2026organization} \\
\midrule
\multirow{2}{*}{\problemdomainbadge{problemgeo}{Geophysics}} & Ice shelf flow laws & \citealp{wang2025ice} \\
& Catchment properties & unpublished expert manuscript \\
\midrule
\multirow{3}{*}{\problemdomainbadge{problemplasma}{Plasma Physics}} & ARC tokamak & \citealp{howard2026arc} \\
& Negative triangularity core-only & \citealp{thome2026negative} \\
& Negative triangularity edge & \citealp{thome2026negative} \\
\midrule
\multirow{7}{*}{\problemdomainbadge{problemastro}{Astrophysics}} & Single galaxy & \citealp{liao2026onegalaxy} \\
& Galaxy environment & \citealp{sims2026environments} \\
& Gravitational wave & \citealp{tillman2026gravitational} \\
& Milky Way & \citealp{garcia2025profiles} \\
& Baryonic feedback & \citealp{garcia2025profiles} \\
& WDM merge tree & \citealp{leisher2025mergertree} \\
& WDM population & \citealp{costanza2025warm} \\
\midrule
\multirow{3}{*}{\problemdomainbadge{problemcs}{Computer Science}} & Stated belief shift & \citealp{geng2025accumulating} \\
& Subliminal learning & \citealp{cloud2025subliminal} \\
& Circuit discovery & \citealp{mueller2025mib} \\
\midrule
\problemdomainbadge{problemchem}{Chemistry} & Reaction discovery & \citealp{zhao2022yarp} \\
\bottomrule
\end{tabular}
\caption{\textbf{Problems and sources in \ourbenchmark.}}
\label{tab:problem_overview}
\end{table}

\end{document}